\documentclass{article}

\usepackage{arxiv}
\PassOptionsToPackage{numbers, compress}{natbib}
\usepackage[numbers]{natbib}

\usepackage{amsmath,amsfonts,bm}

\def\eqref#1{equation~\ref{#1}}
\def\1{\bm{1}}

\DeclareMathAlphabet{\mathsfit}{\encodingdefault}{\sfdefault}{m}{sl}
\SetMathAlphabet{\mathsfit}{bold}{\encodingdefault}{\sfdefault}{bx}{n}

\newcommand{\R}{\mathbb{R}}

\usepackage{amsthm}  % for theorem-like environments
\newtheorem{theorem}{Theorem}[section]

\newtheorem{lemma}[theorem]{Lemma}

\newtheorem{definition}[theorem]{Definition}
\newtheorem{assumption}[theorem]{Assumption}

\usepackage{geometry}
\usepackage[utf8]{inputenc} % allow utf-8 input
\usepackage[T1]{fontenc}    % use 8-bit T1 fonts
\usepackage{hyperref}       % hyperlinks
\usepackage{url}            % simple URL typesetting
\usepackage{booktabs}       % professional-quality tables
\usepackage{nicefrac}       % compact symbols for 1/2, etc.
\usepackage{microtype}      % microtypography
\usepackage{xcolor}         % colors
\usepackage{tikz}
\usepackage{threeparttable}
\usepackage{adjustbox}
\usepackage{comment}

\usepackage{array}
\usepackage{makecell}
\usepackage{multirow}
\usepackage{graphicx}
\usepackage{subcaption}
\usepackage{comment}
\usepackage{wrapfig}	
\usepackage{tabularx}
\usepackage{caption}
\usepackage{booktabs}
\usepackage[tableposition=top]{caption}
\usepackage{float}
\floatstyle{plaintop}
\restylefloat{table}
\usepackage{makecell}
\usepackage{titlesec}
\titlespacing\section{1pt}{1pt}{1pt}
\titlespacing\subsection{1pt}{1pt}{1pt}
\titlespacing\subsubsection{1pt}{1pt}{1pt}
\titlespacing\paragraph{1pt}{1pt}{1pt}
\usepackage[capitalize,noabbrev]
{cleveref}

\newcommand{\mypara}[1]{\vspace{4pt}\noindent\textbf{#1}}

\usepackage{xspace}

\makeatletter
\DeclareRobustCommand\onedot{\futurelet\@let@token\@onedot}
\def\@onedot{\ifx\@let@token.\else.\null\fi\xspace}

\makeatother

\usepackage{enumitem}
\usepackage[titletoc,toc,page]{appendix}

\title{Implicit Causal Representation Learning via Switchable Mechanisms}

 \author{%
	   Shayan Shirahmad Gale Bagi \\
	   Department of Electrical and Computer Engineering \\
	   University of Waterloo \\
	   Waterloo, Canada \\
	   \texttt{sshirahm@uwaterloo.ca} \\
	   \And
	   Zahra Gharaee \\
	   Department of Systems Design Engineering \\
	   University of Waterloo \\
	   Waterloo, Canada \\
	   \texttt{zahra.gharaee@uwaterloo.ca} \\
	   \AND
	   Oliver Schulte \\
	   School of Computing Science \\
	   Simon Fraser University \\
	   Burnaby, Canada \\
	   \texttt{oschulte@cs.sfu.ca} \\
	   \And
	   Mark Crowley \\
	   Department of Electrical and Computer Engineering \\
	   University of Waterloo \\
	   Waterloo, Canada \\
	   \texttt{mark.crowley@uwaterloo.ca} 
	 }

\date{}

\renewcommand{\headeright}{}
\renewcommand{\undertitle}{}
\renewcommand{\shorttitle}{ICLR-SM}

\hypersetup{
pdftitle={SoftCD arxiv},
pdfsubject={q-bio.NC, q-bio.QM},
pdfauthor={Shayan Shirahmad},
pdfkeywords={Motion forecasting, Causality, Domain Adaptation},
}

\begin{document}
\maketitle

\begin{abstract}
Learning causal representations from observational and interventional data in the absence of known ground-truth graph structures necessitates implicit latent causal representation learning. Implicit learning of causal mechanisms typically involves two categories of interventional data: hard and soft interventions. In real-world scenarios, soft interventions are often more realistic than hard interventions, as the latter require fully controlled environments. Unlike hard interventions, which directly force changes in a causal variable, soft interventions exert influence indirectly by affecting the causal mechanism. However, the subtlety of soft interventions impose several challenges for learning causal models. One challenge is that soft intervention's effects are ambiguous, since parental relations remain intact. In this paper, we tackle the challenges of learning causal models using soft interventions while retaining implicit modeling. We propose \mbox{ICLR-SM}, which models the effects of soft interventions by employing a \textit{causal mechanism switch variable} designed to toggle between different causal mechanisms. In our experiments, we consistently observe improved learning of identifiable, causal representations, compared to baseline approaches.

\end{abstract}

% keywords can be removed

\section{Introduction}
\label{sec:intro}
\begin{figure*}[!h]
%\begin{wrapfigure}{r}{0.5\textwidth}
	\centering    
	\includegraphics[width=0.6\textwidth]{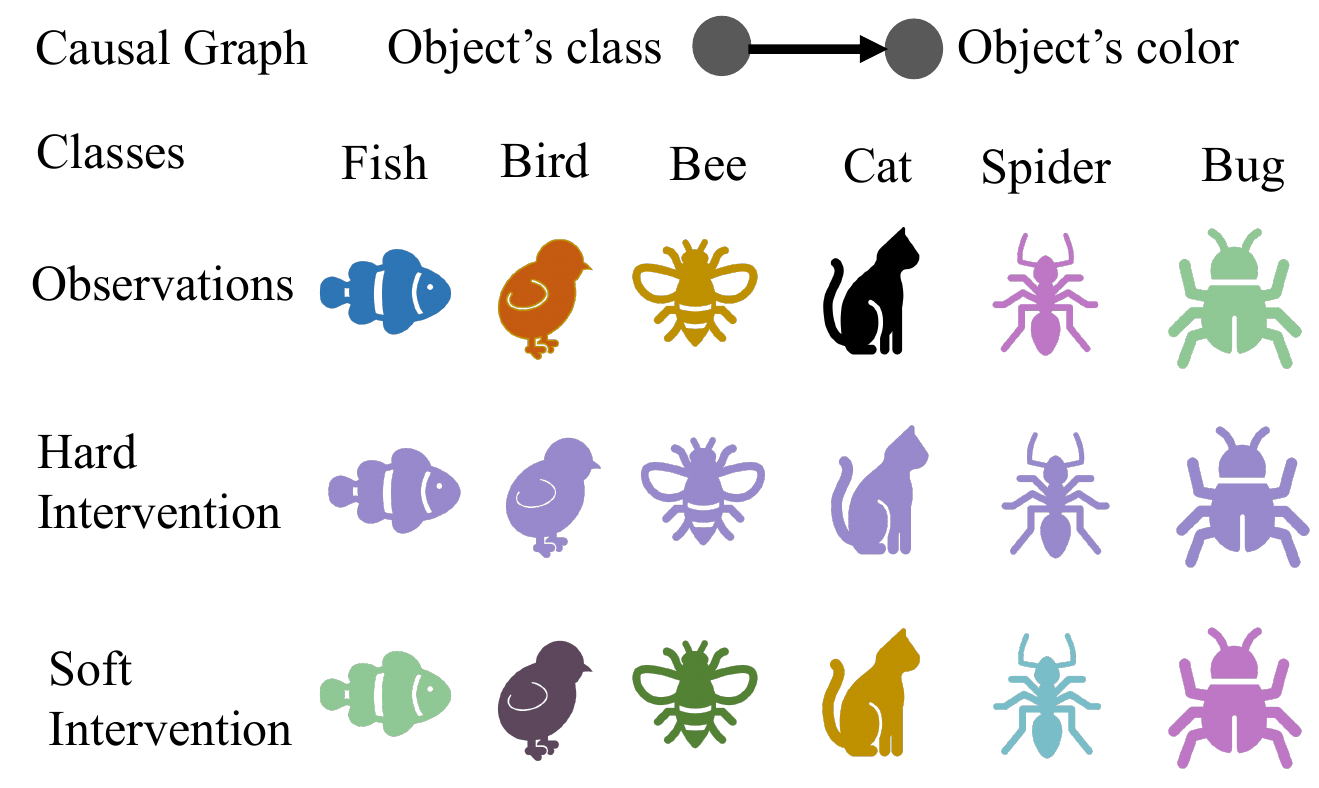}
	\caption{Difference between hard interventions and soft interventions: As seen in the middle row, hard interventions sever connections with parents. Therefore, an object's class cannot have any effect on the object's color when we intervene on color. On the other hand, soft interventions, as shown in the bottom row, allow for such effects.}
	\label{fig:hard_vs_soft_example}
%\end{wrapfigure}
\end{figure*}

One of the long-standing challenges in causal representation learning is how to recover the ground-truth causal graph of a system solely from observations. Termed the \textit{identifiability of causal models} problem, this endeavor is crucial. 
Without achieving identifiability, we risk erroneously attributing causal relationships to learned representations. Furthermore, statistical models can masquerade as Directed Acyclic Graphs (DAGs) where edges lack causal significance, further complicating our pursuit.

When considering the challenge of identifying causal models, it is known that the Markov condition in graphs is insufficient for this task \cite{causality_ML}. Thus, without additional assumptions or data, we find ourselves limited to learning only a \textit{Markov Equivalence Class} (MEC) of the causal model.
Existing works have made different assumptions about availability of ground-truth causal variables labels \cite{causalVAE}, model parameters \cite{interventional_survey}, availability of paired interventional data \cite{ILCM, causal3dident}, and availability of intervention targets \cite{citris} to ensure identifiability of causal models. 

Interventional data are usually obtained through \textit{soft} or \textit{hard} interventions. Hard interventions usually involve controlled experiments and they severe the connection of an intervened variable with its parents \cite{causaljudea}. In terms of Structural Causal Models (SCM), hard interventions set the causal mechanism relating a causal variable to its parents, to a constant. Due to ethical or safety reasons, it may not be possible to perform hard interventions in many real-world applications. On the other hand, the effects of soft interventions are more subtle since parent variables can still affect their children. These effects can be modeled by a change in the set of parents, the causal mechanisms, and the exogenous variables \cite{soft_transportability}. Consequently, hard interventions can also be seen as a special case of soft interventions where the causal mechanism is set to a constant. Illustrated in Figure \ref{fig:hard_vs_soft_example}, a prominent challenge in causal representation learning lies in dealing with the ambiguity surrounding the effects of soft interventions. The observed alterations in object colors fail to distinctly elucidate whether they stem from parental influences or the applied interventions. 

Additionally, a lack of comprehension regarding causal graphs can pose significant challenges in causal representation learning. In certain applications, the causal graph can be constructed using domain knowledge, allowing us to subsequently learn the causal variables \cite{GCRL, csg, causalmotion}. However, this is not universally applicable, necessitating the direct learning of the causal graph itself. In a Variational AutoEncoder (VAE) framework, there are generally two approaches for causal representation learning: Explicit Latent Causal Models (ELCMs) \cite{causalVAE, interventional_survey, citris, CAE, DAG-GNN, sparse_disentanglement} and Implicit Latent Causal Models (ILCMs) \cite{ILCM}.

In ELCMs, the latents are the causal variables and the adjacency matrix of the causal graph is parameterized and integrated into the prior of the latents such that the prior of latents is factorized according to the Causal Markov Condition \cite{causal_survey}. This approach to causal representation learning is highly susceptible to becoming stuck in local minima as it is hard to learn representations without knowing the graph, and it is hard to learn the graph without knowing the representations. ILCMs \cite{ILCM} were introduced to circumvent this ``chicken-and-egg'' problem by using \textit{solution functions}, which can implicitly model edges in the causal graph rather than explicitly modeling the entire adjacency matrix of the causal model. In ILCMs \textit{the latents are the exogenous variables} and the there is no explicit parameterization for the graph. 

In implicit causal representation learning, the task involves recovering the exogenous variables $\mathcal{E}$ from observed variables $\mathcal{X}$ and learning solution functions. In \cite{ILCM}, interventions are assumed to be hard, but this is often unrealistic and does not align with real-world problems. \textbf{In this paper, we propose a novel approach for Implicit Causal Representation Learning via Switchable Mechanisms (ICRL-SM).} We will introduce the \textit{causal mechanism switch variable} as a way of modelling the effect of soft interventions and identifying the causal variables. 

Our experiments with both synthetic and large real-world datasets demonstrate the efficacy of the proposed method in identifying causal variables and suggesting promising future directions for implicit causal representation learning. We validate the theoretical soundness of our approach using synthetic datasets, while also showcasing its practical performance on real-world datasets. 

Our key contributions can be summarized as follows:
\begin{itemize}
	\item A novel approach for implicit causal representation learning with soft interventions.
	\item Employing causal mechanisms switch variable to model the effect of soft interventions.
	\item Theory for identifiability up to reparameterization from soft interventions.

\end{itemize}

\section{Related Work}
\label{gen_inst}
Causal representation learning has recently garnered significant attention \cite{causal_survey, crl_survey}. The primary challenge in this problem lies in achieving identifiability beyond the Markov equivalence class~\cite{causality_ML}. Solely relying on observational data necessitates additional assumptions regarding causal mechanisms, decoders, latent structure, and the availability of interventional data~\cite{causalVAE,interventional_survey,sparse_disentanglement,icarl, DEAR, markovblanket, sparsity_MIS, acyclic_constraint, maximal_ancestral_Graphs}. 
Recent works have focused on identifying causal models from collected interventional data instead of making strong assumptions about functions of the causal model. Interventional data facilitates identifiability based on relatively weak assumptions~\cite{interventional_survey,ILCM,cooper2013causal,causaldiscrepancy,cauca}. This type of data can be further categorized based on whether it involves soft or hard interventions, and whether the manipulated variables are observed and specified or latent. Our focus in this paper is on examining soft interventions, encompassing both observed and unobserved variables.
\begin{table*}[!h]
    \caption{Comparison of proposed method with other recent related work on causal learning from interventional data}
    \centering
    %\resizebox{16cm}{!}{
   \begin{small}
    \begin{tabular}{p{2.2cm}p{1.3cm}p{3.2cm}p{1.5cm}p{1.2cm}p{4.3cm}}
    	\toprule
          & \textbf{Causal} & \textbf{Mixing} &  & \textbf{Learning} &  \\
          
         \textbf{Methods} & \textbf{Mechanism} & \textbf{Function} & \textbf{Intervention} & \textbf{Method} & \textbf{Identifiability} \\
         \midrule
         % CausalVAE \cite{causalVAE} & Linear & Invertible & N/A & Explicit & Affine \\
         CausalDisc.\cite{causaldiscrepancy} %(unpaired data)
         & Nonlinear & Full row rank polynomial & Soft & Explicit & Permutation \& Affine \\
         CauCA \cite{cauca} %(unpaired)
         & Nonlinear & Diffeomorphism & Soft & Explicit & Different based on assumptions \\ 
         Linear-CD \cite{intervention_linear_cd} %(unpaired)
         & Linear & Linear & Hard & Explicit & Permutation \\
         Scale-I \cite{interventions_score_based} %(not NN based)
         & Nonlinear & Linear & Hard/Soft & Explicit & Scale/Mixed \\ 
         ILCM \cite{ILCM} & Nonlinear & Diffeomorphism & Hard & Implicit & Permutation \& reparameterization \\
         dVAE \cite{dvae} & Nonlinear & Diffeomorphism & Hard & Implicit & Permutation \& reparameterization \\
         \textbf{ICRL-SM (ours)} & Nonlinear & Diffeomorphism & Soft & Implicit & Reparameterization \\
    \end{tabular}
    \end{small}   
    %}
    \label{tab:compare_assum}
\end{table*}

\subsection{Explicit models vs. Implicit models}
Table \ref{tab:compare_assum} presents a comparison of the assumptions and identifiability results between our proposed theory and other related works on causal representation learning with interventions.
In causal representation learning with interventions, one approach assumes a given causal graph and concentrates on identifying causal mechanisms and mixing functions. For instance, Causal Component Analysis (CauCA)~\cite{cauca} explores soft interventions with a known graph. Alternatively, when the graph is not provided, explicit models seek to reconstruct it from interventional data \cite{cooper2013causal, citris}, potentially resulting in a chicken-and-egg problem in causal representation learning \cite{ILCM}.

Current methods face the challenge of simultaneously learning the causal graph and other network parameters, especially in the absence of information about causal variables or the graph. Addressing these challenges, \cite{ILCM} recently introduced ILCM, which performs {\em implicit} causal representation learning exclusively using \textit{hard} intervention data. In contrast, our approach introduces a novel method for learning an implicit model from \textit{soft} interventions. \cite{ILCM} describes methods for extracting a causal graph from a learned implicit model, which could be applied to our method as well.

In our experiments, we will compare our method with ILCM and disentangled VAE (dVAE \cite{dvae}), given their implicit nature and similar experimental settings and assumptions. Additionally, to showcase the superiority of our method over explicit models, we will employ explicit causal model discovery methods like ENCO \cite{enco} and DDS \cite{dds}, in conjunction with various variants of $\beta$-VAE~\cite{beta-VAE}.

\subsection{Hard interventions vs Soft interventions}
The identification of explicit causal models from hard interventions has been extensively explored.~\cite{intervention_linear_cd} investigate causal disentanglement in linear causal models with linear mixing functions under hard interventions. Similarly,~\cite{intervention_linear_nonlinearmixing} focus on identifying causal models with linear causal mechanisms and nonlinear mixing functions, also utilizing hard interventions. In a more general setting with non-parametric causal mechanisms and mixing functions,~\cite{unknownI_nonparamteric} examine the identifiability of causal models, utilizing multi-environment data from unknown interventions. Similarly, \cite{GCRL} explore identifiability of causal models using multi-environment data from unknown interventions.~\cite{interventions_score_based} investigate the identifiability of causal models with nonlinear causal mechanisms and linear mixing functions, considering both hard and soft interventions.

Recent work has expanded the concept of explicit hard interventions to include soft interventions. In their study, \cite{causaldiscrepancy} address the identification of causal models from soft interventions, leveraging the sparsity of the adjacency matrix as an inductive bias. However, when dealing with implicit models, soft interventions introduce new complexities. Identifiability becomes more challenging, as the causal effect of variables on observed variables is less apparent. This ambiguity arises from the dual possibility of effects originating from interventions or influences from parent variables on the causal variables. Moreover, in scenarios where implicit modeling is retained, the absence of knowledge about parent variables further complicates identifiability. While \cite{ILCM} theoretically establishes identifiability for hard interventions, practical experiments involving complex causal models with over 10 variables reveal increased ambiguity and confounding factors. Consequently, model identification becomes less straightforward. 

\section{Methodology}
\label{headings}

\begin{figure}[!h]
    \centering
    \includegraphics[width=0.8\textwidth]{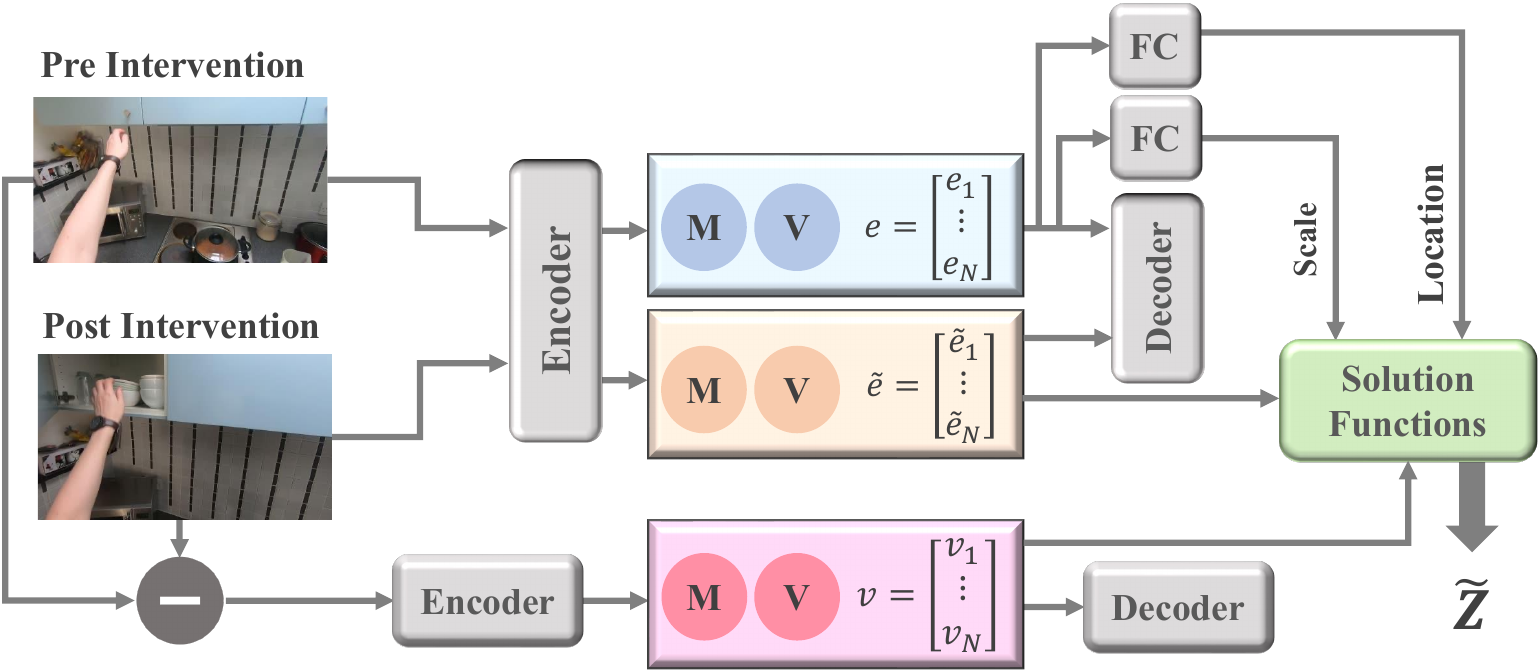}
    \caption{The architecture processes pre-intervention observations $X$, post-intervention observations $\tilde{X}$, and their differences $X - \tilde{X}$ (intervention displacement), encoding them into latent representations. Each encoder outputs the mean (M) and variance (V) of a probability distribution function. By sampling from these distributions, we obtain pre-intervention exogenous variables $E$, post-intervention exogenous variables $\tilde{E}$, and the causal mechanism switch variable $V$. The exogenous variables are derived from the corresponding pre- and post-intervention encodings, while $V$ is obtained from the encoding of the differences between $X$ and $\tilde{X}$. The pre- and post-intervention exogenous variables are then passed through two fully connected (FC) layers, which predict the scale and location parameters. These predicted scale and location parameters, together with the post-intervention exogenous variables and the causal mechanism switch variable, are utilized in the solution function (\ref{eq:solution}) to compute the post-intervention causal variables $\tilde{Z}$. Here, $N$ denotes the total number of causal variables.}
    \label{fig:ICRL-SM}
\end{figure}

\mypara{Notations.}
We would like to note that throughout the paper, we adhere to the notation conventions presented in the Table \ref{tab:notation}.
\subsection{Data Generating Process}

A structural causal model (Definition \ref{def:scm}) is used to understand and describe the relationships between different variables and how they influence each other through causal mechanisms. A \textbf{mixing function}:
\begin{equation}
\label{eq:mix_func}
g(\mathbf{z}) = \mathbf{x},
\end{equation}
which maps a vector of causal values $z$ to observed values $x$. 
The causal variables $Z$ are unobserved and the goal is to infer them from interventional data. For each causal variable, a \textbf{diffeomorphic solution function}:
\begin{equation}
    s_i: \mathcal{E}_i \times \mathcal{E}_{/i} \times \mathcal{V} \rightarrow \mathcal{Z}_i 
    \quad \text{(see Section~\ref{sec:solution_function})}
\end{equation}

%$s_i: \mathcal{E}_i \rightarrow \mathcal{Z}_i$,
deterministically maps a value for exogenous variables $E \in \R^n$ and a value for proposed switch variables $V \in \R^n$ to a value for causal variable $Z_i \in \R$. We will discuss switch variables in details in Section~\ref{sec:switch_variable}. {\em In implicit modeling, we learn the solution functions $s_i$ directly,} rather than defining them through local mechanisms $f_i$. 
 %We write $S$ for the set of all solution functions $s_i \in S$, so $S=[s_1, s_2, ..., s_n]$ for $n$ number of causal variables.

Identifying causal models from data can be complex and is often studied within classes of models such as those identifiable up to affine transformations. 
For example, in the context of nonlinear \textit{Independent Component Analysis (ICA)}, 
the generative process also involves a mixture function $g$ of latent causal variables $Z \in \R^n$, resulting in observations $X \in \R^n$ \cite{sparse_disentanglement, nonlinearICA}. However, a significant distinction between causal representation learning and nonlinear-ICA is that in the former, the causal variables $Z$ may have complex dependencies.
Our objective in this paper is to recover $E$ from $X$ and eventually map $E$ to $Z$ using solution functions.

Identifying a causal model from observational data is not trivial and requires assumptions on the parameters of the model \cite{interventional_survey}. Adding information about interventions in addition to observations, helps to identify causal variables  by exhibiting the effect of changing a causal variable on the observed variables. An interventional data point $(x,\Tilde{x}, i)$ includes the pre-intervention observation $x$, the post-intervention observation $\Tilde{x}$, and intervention target $i \in I$ where $I \subset \{1, 2, ..., n\}$ is the set of intervention targets selected from the causal variables. The post-intervention data $\Tilde{x}$ is generated by a \textit{soft intervention} that targets one of the causal variables. To achieve identifiability up to reparametrization, we rely on a series of assumptions within the data generation process, outlined as follows:

\begin{assumption}(Data generating assumptions)
	\label{assumption:ic}
	%\leavevmode\\
	\begin{itemize}%[noitemsep]
	\item \textbf{Atomic Interventions:} For every sample $(x, \Tilde{x}, i)$, only one causal variable is targeted by an intervention. 
    \item \textbf{Intervention Targets Set:} All interventions are observed, thus $I=\{1,2,...,n\}$.
	\item \textbf{Known Targets:} Targets of soft interventions are known.
	\item \textbf{Post-intervention Exogenous Variables:} The exogenous variables' values change only for the corresponding intervened causal variable, while the others maintain their pre-intervention values, thus $e_i \neq \Tilde{e}_i$ if $i \in I$ ,and $e_i = \Tilde{e}_i$ otherwise.
	\item \textbf{Sufficient Action Variability:} Soft interventions alter causal mechanisms to introduce sufficient action variability \cite{sparse_disentanglement}. These interventions should modify causal mechanisms to ensure non-overlapping conditional distributions of causal variables (refer to Figure \ref{fig:proof}).
	\item \textbf{Diffeomorphic decoder and causal mechanisms:} Diffeomorphism guarantees no information loss and avoids abrupt changes in the function's image.
	\end{itemize}
\end{assumption}
The \textbf{known targets} assumption can be relaxed in applications where such data is not available and the same procedure in \cite{ILCM} can be used to infer the intervention targets. In fact, in our real-world experiments, intervention targets are not available and based on the nature of the datasets, we hypothesize our causal variables to be object attributes and actions to be intervention targets.

\begin{comment}
	Without these assumptions, observed changes in $\tilde{x}$ become ambiguous, as multiple interventions or alterations in exogenous variables may result in overlapping effects on the observed variables. For instance, when examining the combined impact of both heater and lights on room temperature, changing them simultaneously would obscure the precise influence of each on the room temperature. 
	Consequently, the data generation process for an interventional data point $(x,\Tilde{x}, i)$ can be defined as follows:
	\[
	\begin{minipage}[t]{0.45\textwidth}
		\textit{Pre-intervention data:}
		\[
		\left\{
		\begin{aligned}
			&\forall j \quad e_j \sim p(\mathcal{E}_j) \\
			&\forall j \quad z_j = s_j(e_j; e_{/j})
		\end{aligned}
		\right. \quad \rightarrow x = g(z),
		\]
	\end{minipage}
	\hfill
	\begin{minipage}[t]{0.45\textwidth}
		\textit{Post-intervention data:}
		\[
		\left\{
		\begin{aligned}
			&\text{$i$} \quad \quad \quad \tilde{e}_i \sim p(\tilde{\mathcal{E}}_i) \\
			&\forall j \neq i \quad \tilde{e}_j \sim p(\mathcal{E}_j) \\
			&\text{$i$} \quad \quad \quad \tilde{z}_i = \tilde{s}_i(\tilde{e}_i; \tilde{e}_{/i}) \\
			&\forall j \neq i \quad \tilde{z}_j = s_j(\tilde{e}_j; \tilde{e}_{/j}) 
		\end{aligned}
		\right. \quad \rightarrow \tilde{x} = g(\tilde{z})
		\]
	\end{minipage}
	\]
\end{comment}

\subsection{Causal Mechanisms Switch Variable}
\label{sec:switch_variable}
\begin{figure}[!h]
	\centering
	\includegraphics[width=0.5\textwidth]{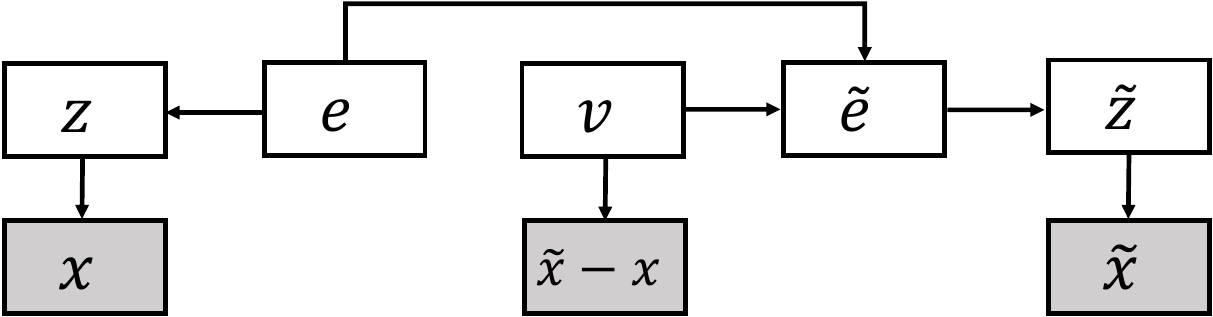}
	\caption{Generative model}
	\label{fig:gen-model}
\end{figure}

The major difference of soft intervention with hard intervention is that post-intervention causal variable $\Tilde{Z_i}$ is no longer disconnected from its parents and its post-intervention solution function $\Tilde{s}_i$ is affected by the intervention. This is why identifying the causal mechanisms is more difficult for soft interventions. Soft intervention data yield fewer constraints on the causal graph structure than hard intervention data. For more details refer to string diagrams of soft and hard interventions depicted in Figure~\ref{fig:hard_soft_int}. Figure~\ref{fig:gen-model} shows our main generative model. It includes a data augmentation step that adds the intervention displacement $\Tilde{x}-x$ as an observed feature that directly represents the effect of a soft intervention in observation space.

\mypara{Augmented implicit causal model.}
To model the effect of soft interventions, we introduce the causal mechanism switch variable $V$ \cite{causality_ML}. By leveraging $V$, we can effectively switch to the pre-intervention causal mechanisms within post-intervention data. This facilitates the model's ability to solely focus on discerning alterations in the intrinsic characteristics of each causal variable. These changes are encapsulated within their respective exogenous variables, aiding the model in learning the causal relationships more accurately. We propose to use a modulated form of $V$ to model the soft intervention effects on each causal variable with a nonlinear function $h_i: \mathcal{V} \rightarrow \R$ such that:

\begin{equation}
\label{eq:aicm}
i \in I, \: \Tilde{z_i} = \Tilde{s}_i(\Tilde{e}_i;\Tilde{e}_{/i}) = s_i(\Tilde{e}_i;e_{/i}, h_i(v))
\end{equation}

As the parental set for each causal variable is not known, we have to use a modulated form of $V$ in every causal variable's solution function and the inclusion of $h_i(v)$ enables the model to encompass variations in the parental sets of all causal variables in $V$. Therefore, there is a switch variable $V_i \in \R$ for each causal variable $Z_i$. Adding switch variables to solution functions leads to the concept of an \textit{augmented implicit causal model}. 

\begin{definition}(Augmented Implicit Causal Models)
	\label{def:ascm}
	An Augmented Implicit Causal Models (AICMs) is defined as $\mathcal{A}=(\mathcal{S}, \mathcal{Z}, \mathcal{E}, \mathcal{V})$ where $V \in \R^n$ is the {\em causal mechanism switch variable}, which models the effect of soft interventions on solution functions in $\mathcal{S}$ as shown by (\ref{eq:aicm}).
\end{definition}
The usage of $V$ in soft interventions is analogous to augmented networks in \cite{causality} mainly designed for hard interventions. Pearl~\cite{causality} even foresaw this possibility by saying:~
\textquotedbl{}One advantage of the augmented network representation is that it is applicable to any change in the functional relationship $f_i$ and not merely to the replacement of $f_i$ by a constant.\textquotedbl{}

By using Taylor's expansion, we can expand the solution functions as follows:
\begin{align}
	\label{eq:taylor}
	\begin{split}
		\resizebox{\linewidth}{!}{$s_i(\tilde{e}_i; e_{/i}, h_i(v)) = s_i(\tilde{e}_i; e_{/i}, h_i(v_0)) + \sum_{n=1}^\infty \frac{1}{n!} \left( \frac{\partial^n s_i}{\partial h_i^n} \bigg|_{h_i=h_i(v_0)} (h_i(v) - h_i(v_0))^n \right) = s_i(\tilde{e}_i; e_{/i}, h_i(v_0)) + R_i $}
	\end{split}
\end{align}
where we'll use $R_i$ as a short-hand of second term in (\ref{eq:taylor}). We define the \textbf{separable dependence} property for solution functions as:
\begin{equation}
	\exists h_i(v_0): s_i(\Tilde{e}_i; e_{/i}, h_i(v_0)) = s_i(\Tilde{e}_i; e_{/i}).
\end{equation}

An example of such a scenario could be in location-scale noise models such that:
\begin{equation}
	s_i(\Tilde{e_i};e_{/i}, h_i(v)) = \Tilde{e_i}+loc(e_{/i}) + h_i(v) = \Tilde{e_i} + loc(e_{/i}) + v^2 + v,
\end{equation}
where $v_0$ would be zero. Ignoring argument $h_i(v_0)$ in this example means settings it to zero. More generally, it could involve setting it to any other value.

By assuming the separable dependence property, we can write the solution function in (\ref{eq:taylor}) as:
\begin{align}
	\label{eq:taylor_simple}
	\begin{split}
		s_i(\Tilde{e}_i; e_{/i}, h_i(v)) = s_i(\Tilde{e}_i; e_{/i}) + R_i = s_i(\Tilde{e}_i; e_{/i}) + \textit{soft intervention effect}
	\end{split}
\end{align}
As a result, we can switch to pre-intervention solution functions. Subsequently, by modeling soft intervention effects using $h_i(v)$, we can recover pre-intervention solution functions. During inference, we simply disregard the $h_i(v)$ term in the solution functions. Nonetheless, it is possible to train the prior $p(v)$ to ensure that the separable dependence property is maintained for pre-intervention data.

\mypara{Observability of switch variable.}
The intuition behind using $V$ is to separate the effect of soft intervention on $\Tilde{Z}_i$ into:
\begin{enumerate}[noitemsep]
\item The effect on causal mechanisms and parents, 
\item The effect on exogenous variable $E_i$. 
\end{enumerate}
For example, we can say that causal variables in images of objects are the objects' attributes such as shape, color, and size, and performing actions like "Fold" change these attributes. Furthermore, it can be asserted that the camera angle within a given image may influence the shape of the object. If the images were generated from a hard intervention, the camera angle remains fixed between pre and post intervention. However, the camera angle changes along with the performed actions indicating that the interventions are soft. In this case, if we had a knowledge of how the camera angle affects the attributes of objects, then we could separate the effect of soft intervention. In other words, if $V$ is observed, then we can extract the effect of the intervention that we are interested in (i.e., the effect on the causal variable itself). For more details, refer to Figure~\ref{fig:causal_triplet}.

Lacking an understanding of how soft intervention influences the causal model, a more complex model becomes necessary. Consequently, the term $R_i$ in (\ref{eq:taylor}) would involve a higher order of $h_i(V)$. Therefore, we assume the observability of $V$.  
\begin{assumption}(Observability of $V$)
	\label{assumption:observev} Given an intervention sample $(x,\Tilde{x}, i)$ and linear mixing functions, we can approximate the soft intervention effects $R_i$ as follows:
    \begin{equation}
    \label{eq:dz}
    		\Tilde{z}-z \stackrel{\text{(\ref{eq:taylor})}}{=} \Delta{e_i} + R 
    \end{equation}
    \begin{equation}
    \Tilde{x} - x \stackrel{\text{(\ref{eq:mix_func})}}{=} g(\Tilde{z}) - g(z) 
    \stackrel{\text{Linearity}}{=} g(\Tilde{z} - z) \stackrel{\text{(\ref{eq:dz})}}{=} g(\Delta{e_i} + R),
\end{equation}
where $R=[R_0, R_1, ..., R_n ]$ and $n$ is the number of causal variables. $R$ and $\Delta{e_i}$ are the vectors indicating the soft intervention effects and change in the effect of the exogenous variable of the intervened causal variable, respectively. Note that elements of $R$ will be all zero except for the intervened causal variable. Consequently, with linear mixing function and some pre-processing on observed samples (here subtraction), we can observe $R_i$.
	
	Note that for nonlinear mixing functions, augmentations other than subtraction may be used to obtain information about causal mechanism switch variable $V$.
\end{assumption}

Our synthetic data is generated using a linear mixing function, however, the mixing function for the real-world datasets is not necessarily linear. Therefore, we do not observe $V$ from $\Tilde{x} - x$ in the real-world dataset. Nevertheless, our findings suggest that incorporating soft interventions through $V$ leads to superior performance compared to other implicit modeling approaches. Clearly, understanding the impact of soft interventions on the generative system of the dataset would result in improved outcomes.

%\begin{wrapfigure}{r}{0.4\textwidth}
    %\begin{figure}[!h]
	%\centering
	%\includegraphics[width=0.4\textwidth]{figures/main_model.pdf}
	%\caption{Our generative model. $x,\Tilde{x}$ represent low-level observed variables. $e,\Tilde{e}$ represent higher-level latent exogenous variables, and $z,\Tilde{z}$ represent higher-level latent causal variables. $v$ represents the causal mechanism switch variable. }
	%\label{fig:gen-model}
    %\end{figure}
%\end{wrapfigure}

%\begin{figure}[!h]
%    \centering
%    \begin{subfigure}[b]{0.55\textwidth}
%        \centering
%        \includegraphics[width=\textwidth]{figures/soft_ilcm_arc.pdf}
%        \caption{General overview of ICRL-SM}
%        \label{fig:ICRL-SM}
%    \end{subfigure}%
%    \hfill
%    %\hspace{0.5em}
%    \begin{subfigure}[b]{0.35\textwidth}
%        \centering
%        \includegraphics[width=\textwidth]{figures/main_model.pdf}
%        \caption{Generative model}
%        \label{fig:gen-model}
%    \end{subfigure}
%    %\caption{Comparison of figures}
%    \label{fig:generative_model}
%\end{figure}

\subsection{Identifiability Theorem for Implicit SCMs with Soft Interventions}
In this paper, our focus lies in identifying the causal variables up to reparameterization through soft interventions. We first define identifiability up to reparameterization (Definition~\ref{def:equivalence}) and subsequently introduce the identifiability theorem~\ref{theory:ident}. The proof of theorem is extensive and is available in full in Appendix~\ref{append:proof}.

We establish identifiability up to reparameterization, allowing for the mapping of causal variables in $\mathcal{Z}$ and $\mathcal{Z}'$ between two Latent Causal Models ($\mathcal{M}$ and $\mathcal{M}'$) through component-wise transformations (Definition \ref{def:component-wise}). Given our implicit modeling approach, lacking knowledge of the causal graph, we include all exogenous variables in the solution functions, as depicted in (\ref{eq:aicm}). Notably, \textbf{the causal graph remains unaltered during learning}.

To illustrate, we contrast hard interventions, which neglect parent influences, with soft interventions that acknowledge parental effects in a simple example. Consider a basic causal model $Z_1 \rightarrow Z_2$ alongside a location-scale noise model \cite{lsnm} for the solution function, given by:
\begin{equation}
\tilde{z}_2 = \frac{\tilde{e}_2 - \widetilde{loc}(e_1)}{\widetilde{scale}(e_1)}.
\end{equation}
The distribution $p(\tilde{Z}_2)$ mean is:
\begin{equation}
	\frac{1}{\widetilde{scale}(e_1)} \times \text{mean}(\tilde{E}_2) - \frac{\widetilde{loc}(e_1)}{\widetilde{scale}(e_1)}.
\end{equation}

In the context of hard interventions, we can assume $p(\tilde{Z}_2|Z_1=z_1) = p(\tilde{Z}_2)=N(0,1)$ as there are no parental effects. Consequently, the location and scale networks within the solution function tend to dampen parental effects, given the absence of parental influence in the ground-truth data. Contrarily, soft interventions exhibit parental influence in the ground-truth data, thus $p(\tilde{Z}_2|Z_1=z_1) \neq N(0,1)$. Due to the lack of parental knowledge in implicit modeling, we model:
\begin{equation}
    p(\tilde{Z}_2|Z_1=z_1)=p(\tilde{Z}_2|E_2=e_2),
\end{equation}
where $E_2$ is a known parent of $\tilde{Z}_2$. Consequently, parental effects are propagated to $E_i$ (the corresponding exogenous variable of each causal variable), violating identifiability up to reparameterization. By leveraging $V$, we allow parental effects to propagate to $V$ instead of $E_i$.

\begin{definition}(Equivalence up to component-wise reparameterization)
	\label{def:equivalence}
	Let $\mathcal{M}=(\mathcal{A}, \mathcal{X}, g, \mathcal{I})$ and $\mathcal{M}'=(\mathcal{A}', \mathcal{X}, g', \mathcal{I})$ be two Latent Causal Models  (LCM) based on AICMs $\mathcal{A},\mathcal{A}'$ with shared observation space $\mathcal{X}$, shared intervention targets $\mathcal{I}$, and respective decoders $g$ and $g'$.
	We say that $\mathcal{M}$ and $\mathcal{M}'$ are equivalent up to component-wise reparameterization $\mathcal{M} \sim_r \mathcal{M}'$ if there exists a component-wise transformation (Definition \ref{def:component-wise}) $\phi_{\mathcal{Z}}$ from the causal variables in $\mathcal{Z}$ to the causal variables in $\mathcal{Z}'$ and a component-wise transformation $\phi_{\mathcal{E}}$ between $\mathcal{E}$ and $\mathcal{E}'$ such that:
	
	\begin{itemize}[noitemsep]
	\item  Indices are preserved (i.e., $\phi_i(z_i)=z'_{i}$ and $\phi_i(e_i)=e'_{i}$). Corresponding edges are preserved (i.e., $Z_i \rightarrow Z_j$ holds in $G$ iff $Z'_{i} \rightarrow Z'_{j}$ holds in $G'$. Edges $E_i \rightarrow Z_i$ should be preserved as well.)
	\item  The exogenous transformation preserves the probability measure on exogenous variables 
	$p_{\mathcal{E}'} = (\phi_{\mathcal{E}})_*p_{\mathcal{E}}$ (Definition \ref{def:pushmeasure}).
	\item  The causal transformation preserves the probability measure on causal variables  $p_{\mathcal{Z}'} = (\phi_{\mathcal{Z}})_*p_{\mathcal{Z}}$ (Definition \ref{def:pushmeasure}). 
	\end{itemize}
\end{definition}

\begin{theorem}(Identifiability of latent causal models.)
	\label{theory:ident}
	Let $\mathcal{M}=(\mathcal{A}, \mathcal{X}, g, \mathcal{I})$ and $\mathcal{M}'=(\mathcal{A}', \mathcal{X}, g', \mathcal{I})$ be two LCMs with shared observation space $\mathcal{X}$ and shared intervention targets $\mathcal{I}$. Suppose the following conditions are satisfied:
	\begin{enumerate}[noitemsep]
	\item  Data generating assumptions explained in Assumption \ref{assumption:ic}.
	\item  Soft interventions satisfy Assumption \ref{assumption:observev}.
	\item  The causal and exogenous variables are real-valued. 
	\item  The causal and exogenous variables follow a multivariate normal distribution. 
	\end{enumerate}
	Then the following statements are equivalent:
    \begin{enumerate}[noitemsep]
        \item Two LCMs $\mathcal{M}$ and $\mathcal{M}'$ assign the same likelihood to interventional and observational data i.e.:
     \begin{equation}
         p^{\mathcal{X}, \mathcal{I}}_{\mathcal{M}}(x,\Tilde{x}, i) = p^{\mathcal{X}, \mathcal{I}}_{\mathcal{M}'}(x, \Tilde{x}, i)
     \end{equation} 
	\item $\mathcal{M}$ and $\mathcal{M}'$ are disentangled, that is $\mathcal{M} \sim_r \mathcal{M}'$ according to Definition \ref{def:equivalence}.
    \end{enumerate}
	
\end{theorem}

%\textcolor{red}{Do you need to assume they assign the maximum likelihood, i.e. one is the ground truth model?} \textcolor{blue}{This is not an assumption, it is one way of defining identifiability itself. This statement basically says that with one dataset (observations x and $\Tilde{x}$), the model is guaranteed to learn a unique set of causal models, i.e., set of equivalent causal models up to reparameterization as defined in \ref{def:equivalence}.}\\

\subsection{Training Objective}
Consequently, there will be three latent variables in ICRL-SM:
\begin{enumerate}[noitemsep]
 \item A causal mechanism switch variable $V$. 
 \item The pre-intervention exogenous variables $E$. 
 \item The post-intervention exogenous variables $\Tilde{E}$.
 \end{enumerate}
 
 As the data log-likelihood $\log p(x, \Tilde{x},\Tilde{x}-x) \equiv \log p(x, \Tilde{x}) $ is intractable, we utilize an ELBO approximation as training objective:
\begin{equation}
	\resizebox{\linewidth}{!}{
		$\begin{aligned}
			\log p(x, \tilde{x}) \geq & E_{q(e,\tilde{e},v|x,\tilde{x})}\Bigl[\log p(x,\tilde{x}|e,\tilde{e},v)\Big] - KLD(q(e,\tilde{e},v|x,\tilde{x})||p(e,\tilde{e},v)) \\
			& = E_{q(v|\tilde{x} - x) \cdot q(e|x) \cdot q(\tilde{e}|\tilde{x})}\Bigl[\log (p(x|e)p(\tilde{x}|\tilde{e})p(\tilde{x} - x|v))\Big] - KLD(q(v|\tilde{x} - x) \cdot q(e|x) \cdot q(\tilde{e}|\tilde{x})||p(\tilde{e}|e, v)p(v)p(e)).
		\end{aligned}$}
\end{equation}
The observations are encoded and decoded independently. The KLD term regularizes the encodings to share the latent {\em intervention model} $p(\Tilde{e}|e, v)p(v)p(e)$ that is shared across all data points. The components of this model can be interpreted as follows:

\begin{enumerate}[noitemsep]
\item {~$p(e)$} is the prior distribution over exogenous variables $e$. 
\item {~$p(v)$} is the prior distribution over switch variables $v$.
\item {~$p(\tilde{e}|e,v)$} is a transition model that shows how the exogeneous variables change as a function of the intervention.
\end{enumerate}

We factorize the posterior with a mean-field approximation:
\begin{equation}
	q(v,e,\Tilde{e}|x,\Tilde{x}) = q(v|\Tilde{x} - x) \cdot q(e|x) \cdot q(\Tilde{e}|\Tilde{x}).
\end{equation}

Following our data generation model (\cref{fig:gen-model}), the reconstruction probability is factorized as:
\begin{equation}
	p(x, \Tilde{x}|e,\Tilde{e},v) =  p(x|e)p(\Tilde{x}|\Tilde{e})p(\Tilde{x} - x|v).
\end{equation}

The prior over latent variables is factorized as:
\begin{equation}
	p(\Tilde{e}, e, v)= p(\Tilde{e}|e, v)p(v)p(e) \quad (\text{\cref{fig:gen-model}}).
\end{equation}

Pre-intervention exogenous variables are mutually independent, hence:
\begin{equation}
	p(e) = \prod_i p(e_i) \quad \text{and} \quad p(v) = \prod_i p(v_i).
\end{equation}

We assume $p(e_i)$ and $p(v_i)$ to be standard Gaussian. Furthermore, as we assume $e_i=\Tilde{e}_i$ for all non-intervened variables, the $p(\Tilde{e}|e,v)$ will be as follows:
\begin{equation}
	\label{eq:prior_tilde_e}
	\begin{aligned}
		p(\Tilde{e}|e,v) = \Pi_{i \notin I} \: \delta(\Tilde{e}_i -e_i)\Pi_{i \in I} \: p(\Tilde{e}_i|e,v) = \Pi_{i \notin I} \: \delta(\Tilde{e}_i -e_i)\Pi_{i \in I} \: p(\Tilde{z}_i|e_i)\left|\frac{\partial \Tilde{z}_i}{\partial \Tilde{e}_i}\right|.
	\end{aligned}
\end{equation}
The last equality in (\ref{eq:prior_tilde_e}) is obtained from the Change of Variable Rule in probability theory, applied to the solution function $\Tilde{z}_i = s_i(\Tilde{e}_i;e_{/i}, h_i(v))$. Furthermore, we write $p(\Tilde{z}_i|e, v)=p(\Tilde{z}_i|e_i)$ since only $e_i$ is a known parent of $\tilde{z}_i$ in implicit modeling. We assume $p(\Tilde{z}_i|e_i)$ to be a Gaussian whose mean is determined by $e_i$.
%Since we use normalizing flows to model the solution function and  \cite{rnvp}, $\Tilde{s_i}$ is . 

\subsection{Solution Functions}
\label{sec:solution_function}
We implement the solution function using a location-scale noise models \cite{lsnm} as also practiced in \cite{ILCM}, which defines an invertible diffeomorphism. For simplicity, in our experiments, we are only going to change the $loc$ network in post-intervention. Therefore, $h_i(v)$ will be used as:
\begin{equation}
	\label{eq:solution}
	\Tilde{z}_i = \Tilde{s}_i(\Tilde{e}_i;e_{/i},h_i(v))=\frac{\Tilde{e}_i - (loc_i(e_{/i}) + h_i(v))}{scale_i(e_{/i})},
\end{equation}
where $loc_i:\R^{n-1} \rightarrow \R$ and $scale_i:\R^{n-1} \rightarrow \R$ are fully connected networks calculating the first and second moments, respectively. 
%Through this solution function, the model implicitly learns which of $e_{/i}$ are ancestral exogenous variables, hence, identifying causal relations. 
%We also obtain $v$ from $\Tilde{x} -x$ to ensure that $v$ should capture the effects of soft intervention, since we assume the changes in $\Tilde{x}$ are due to the soft intervention only.
The general overview of the model is illustrated in Figure~\ref{fig:ICRL-SM}.

\section{Experiments and Results}
\label{others}

\mypara{Recovering Causal Variable.}
We validate our theorem using synthetic datasets to demonstrate its theoretical soundness. In these experiments, we evaluate our proposed model under conditions where all assumptions are met, thus serving as a \textit{proof of concept}. Specifically, we assume a Gaussian data distribution and a linear mixing function to assess the model’s performance. In these experiments we use causal disentanglement to identify the true causal graph from pairs of observations $(x, \Tilde{x}, i)$.

\mypara{Action and Object Inference.}
We test our model on real-world datasets, namely Epic-Kitchen and ProcTHOR from the Causal-Triplet benchmark. These datasets may not meet all assumptions, such as the diffeomorphic decoder and the observability of the switch variable ($V$). For these experiments, we employ a ResNet-based decoder. Given that the data distribution in real-world scenarios is often unknown, it may not adhere to a Gaussian distribution. 

Since ground truth causal variables were not available in the real-world datasets, we used action and object accuracy as proxies to evaluate the effectiveness of our method in uncovering causal variables compared to other approaches. To that end, we employ two classifiers which take in as input the pre and post intervention causal variables to predict the action and object class.

Moreover, we conducted additional experiments designed as an ablation study, the results of which are presented in~\ref{sec:ablation}. All models are trained using the same setting and data with known intervention targets. Note that in our real-world experiments, the intervention targets are unknown, and we have hypothesized them. Furthermore, the results in \cite{ILCM} have indicated that inferring intervention targets is a relatively easier task; A similar procedure in \cite{ILCM} can be used to infer intervention targets in applications where they are unknown.

\subsection{Datasets}
\mypara{Synthetic Dataset.} 
We generate simple synthetic datasets with $\mathcal{X} = \mathcal{Z} = \R^n$. For each value of $n$, we generate ten random DAGs, a random location-scale SCM, then a random dataset from the parameterized SCM. 
%with random DAGs and causal mechanisms, in which 
To generate random DAGs, each edge is sampled in a fixed topological order from a Bernoulli distribution with probability 0.5. The pre-intervention and post-intervention causal variables are obtained as:
\begin{equation}
	z_i = \text{scale}(z_{pa_i})e_i + \text{loc}(z_{pa_i}) \xrightarrow{\text{Soft-Intervention}} \Tilde{z}_i = \text{scale}(z_{pa_i})\Tilde{e}_i + \widetilde{\text{loc}}(z_{pa_i}),
\end{equation}
where the $loc$ and $scale$ networks are changed in post intervention. The pre-intervention $loc$ and post-intervention $\widetilde{loc}$ network weights are initialized with samples drawn from $\mathcal{N}(0,1)$ and $\mathcal{N}(3,1)$, respectively. The $scale$ is constant 1 
%(no effect from parents) 
for both pre-intervention and post-intervention samples. Both $e_i$ and $\Tilde{e_i}$ are sampled from a standard Gaussian. The causal variables are mapped to the data space through a randomly sampled $SO(n)$ rotation. For each dataset, we generate 100,000 training samples, 10,000 validation samples, and 10,000 test samples. 

\mypara{Real-world Dataset.} 
Causal-Triplet datasets tailored for \textit{actionable} counterfactuals \cite{causal_triplet} feature paired images where several global scene properties may vary including camera view and object occlusions. Thus, the images can be viewed as outcomes of soft interventions, wherein actions affect objects alongside subtle alterations. These datasets consist of:
\begin{itemize}[noitemsep]
    \item \textbf{ProcTHOR~}\cite{deitke2022} images obtained from a photo-realistic simulator of embodied agents.
    \item \textbf{Epic-Kitchen~}\cite{epcikitchens} images repurposed from a real-world video dataset of human-object interactions. 
\end{itemize}

The ProcTHOR dataset contains 100\,k images in which 7 types of actions manipulate 24 types of objects in 10\,k distinct ProcTHOR indoor environments. The second dataset consists of 2,632 image pairs, collected under a similar setup from the Epic-Kitchens dataset with 97 actions manipulating 277 objects. Based on the nature of actions in this dataset, the causal variables should represent attributes of objects such as shape and color. As the dataset consists of images we train all methods with ResNet encoder and decoder. For the ProcThor dataset the number of causal variables are 7. For the Epic-Kitchens dataset, we randomly chose 20 actions from the dataset as 97 causal variables will be too complex in a VAE setup.

\subsection{Metrics} 
For the causal disentanglement task, we are going to use the Disentanglement (D), Completeness (C) Informativeness (I) scores (DCI) \cite{dci_scores}. According to \cite{dci_scores}, Disentanglement (D) and Completeness (C) are sufficient to evaluate the correspondence of learned latents and true causal variables. Therefore, we have utilized the following metrics to evaluate our model. 

\mypara{Causal disentanglement.} 
Causal disentanglement score quantifies the degree to which the modeled causal variable $Z_i$ factorises or disentangles the groundtruth causal variable $Z_i^*$. Causal disentanglement $D_i$ for $Z_i$ is calculated as:
\begin{equation}
\label{eq:d_dci}
 D_i=(1 - H_K(P_{i.})) = (1 + \sum_{k=0}^{K-1}P_{ik} \log_K P_{ik}),
\end{equation} 
where $P_{ij}=\frac{R_{ij}}{\sum_{k=0}^{K-1} R_{ik}}$ and $R_{ij}$ denotes the probability of $Z_i$ being important for predicting $Z_j^*$. Total causal disentanglement is the weighted average $\sum_i \rho_iD_i$ where $\rho_i=\frac{\sum_j R_{ij}}{\sum_{ij}R_{ij}}$. 

\mypara{Causal Completeness.}
Causal Completeness quantifies the degree to which each groundtruth causal variable $Z_i^*$ is captured by a single modeled causal variable $Z_i$. Causal completeness is calculated as:
\begin{equation}
\label{eq:c_dci} 
C_j= (1 - H_D(\Tilde{P}_{.j})) = (1 + \sum_{d=0}^{D-1}\Tilde{P}_{dj} \log_D \Tilde{P}_{ij}).
\end{equation}
$D$ (\ref{eq:d_dci}) and $K$ (\ref{eq:c_dci}) are equal to the dimension of $Z$ and $Z^*$ which is $n$.

\mypara{Classification Accuracy.}
DCI~\cite{dci_scores} scores require ground-truth causal variables, which are not available in the causal-triplet datasets. Therefore, in our experiments with real-world datasets, we will use classification accuracy as the evaluation metric for the action inference task. 

\section{Results}
\label{sec:results}
\subsection{Causal Disentanglement}
We generated a dataset for the soft interventions and trained the models of ICRL-SM, ILCM, $\beta$-VAE and dVAE for 10 different seeds, which generated 10 different causal graphs. We selected 4 causal variables to encompass complex causal structures, including forks, chains, and colliders. Table~\ref{tab:D4} displays the Causal Disentanglement and Causal Completeness scores for all models, computed on the test data. 

Note that the proposed model in \cite{dvae} applied an averaging function to the posterior of their causal variables as a substitute to one of their constraints which is not applicable in our setting. Namings used in \cite{dvae} is based on this averaging function. Following \cite{ILCM}, we have also used the naming as disentangled VAE (dVAE).

\begin{table}[!h]
%\begin{minipage}{0.5\textwidth}
	\centering
	\begin{adjustbox}{width=1\textwidth}
		\begin{small}
			\begin{threeparttable}
				\begin{tabular}{cc cccc cccc}
					\toprule
					\multicolumn{2}{c}{\textbf{Graph}} &
					\multicolumn{4}{c}{\textbf{Causal Disentanglement}} & \multicolumn{4}{c}{\textbf{Causal Completeness}} \\
					
					\cmidrule(lr){1-2} \cmidrule(lr){3-6} \cmidrule(lr){7-10}
					
					\textbf{Model} & \textbf{Name} 
					& {$\beta$-VAE~\cite{beta-VAE}} & {dVAE~\cite{dvae}} & {ILCM~\cite{ILCM}}  & \textbf{ICRL-SM (ours)} 
					& {$\beta$-VAE~\cite{beta-VAE}} & {dVAE~\cite{dvae}} & {ILCM~\cite{ILCM}} & \textbf{ICRL-SM (ours)}\\
					\midrule
					% G-1
					\multirow{2}{*}{\scalebox{0.5}{
							\begin{tikzpicture}
								\node[draw, circle, fill=red] (A) at (0,0){};
								\node[draw, circle, fill=red] (B) at (1,0){};
								\node[draw, circle, fill=red] (C) at (0,-1){};
								\node[draw, circle, fill=red] (D) at (1,-1){};
								\draw[->, line width=2pt, color=red] (A) -- (C); 
								\draw[->, line width=2pt, color=red] (A) -- (D); 
								\draw[->, line width=2pt, color=red] (B) -- (C);
							\end{tikzpicture}
					}}
					\\
					&G1 &0.38&0.54&0.71&\textbf{0.82}      &0.51 &0.69&0.78&\textbf{0.87}  \\
					% G-2
					\multirow{2}{*}{\scalebox{0.5}{
							\begin{tikzpicture}
								\node[draw, circle, fill=blue] (A) at (0,0){};
								\node[draw, circle, fill=blue] (B) at (1,0){};
								\node[draw, circle, fill=blue] (C) at (0,-1){};
								\node[draw, circle, fill=blue] (D) at (1,-1){};
								\draw[->, line width=2pt, color=blue] (A) -- (B); 
								\draw[->, line width=2pt, color=blue] (C) -- (D); 
							\end{tikzpicture}
					}}
					\\
					&G2 &0.30&0.72&0.75&\textbf{0.83}     &0.49&0.77&0.80&\textbf{0.87}  \\
					% G-5
					\multirow{2}{*}{\scalebox{0.5}{
							\begin{tikzpicture}
								\node[draw, circle, fill=green] (A) at (0,0){};
								\node[draw, circle, fill=green] (B) at (1,0){};
								\node[draw, circle, fill=green] (C) at (0,-1){};
								\node[draw, circle, fill=green] (D) at (1,-1){};
								\draw[->, line width=2pt, color=green] (A) -- (B); 
								\draw[->, line width=2pt, color=green] (A) -- (C); 
								\draw[->, line width=2pt, color=green] (B) -- (C);
								\draw[->, line width=2pt, color=green] (C) -- (D);
							\end{tikzpicture}
					}}
					\\
					&G3 &0.28 &0.51 &0.68 &\textbf{0.98}     &0.49&0.56&0.78&\textbf{0.98}  \\
					% G-6
					\multirow{2}{*}{\scalebox{0.5}{
							\begin{tikzpicture}
								\node[draw, circle, fill=black] (A) at (0,0){};
								\node[draw, circle, fill=black] (B) at (1,0){};
								\node[draw, circle, fill=black] (C) at (0,-1){};
								\node[draw, circle, fill=black] (D) at (1,-1){};
								\draw[->, line width=2pt, color=black] (A) -- (C); 
								\draw[->, line width=2pt, color=black] (A) -- (D); 
								\draw[->, line width=2pt, color=black] (B) -- (D);
							\end{tikzpicture}
					}}
					\\
					&G4 &0.16&0.50&0.65&\textbf{0.68}        &0.38&0.69&0.77&\textbf{0.78}  \\
					% G-7
					\multirow{2}{*}{\scalebox{0.5}{
							\begin{tikzpicture}
								\node[draw, circle, fill=pink] (A) at (0,0){};
								\node[draw, circle, fill=pink] (B) at (1,0){};
								\node[draw, circle, fill=pink] (C) at (0,-1){};
								\node[draw, circle, fill=pink] (D) at (1,-1){};
								\draw[->, line width=2pt, color=pink] (A) -- (B); 
								\draw[->, line width=2pt, color=pink] (A) -- (C); 
								\draw[->, line width=2pt, color=pink] (A) -- (D);
								\draw[->, line width=2pt, color=pink] (B) -- (C);
								\draw[->, line width=2pt, color=pink] (B) -- (D);
								\draw[->, line width=2pt, color=pink] (C) -- (D);
							\end{tikzpicture}
					}}
					\\
					&G5 &0.27&0.44&\textbf{0.53}&0.42          &0.45&0.54&\textbf{0.66}&0.50  \\
					% G-8
					\multirow{2}{*}{\scalebox{0.5}{
							\begin{tikzpicture}
								\node[draw, circle, fill=purple] (A) at (0,0){};
								\node[draw, circle, fill=purple] (B) at (1,0){};
								\node[draw, circle, fill=purple] (C) at (0,-1){};
								\node[draw, circle, fill=purple] (D) at (1,-1){};
								\draw[->, line width=2pt, color=purple] (A) -- (B); 
								\draw[->, line width=2pt, color=purple] (B) -- (C); 
								
							\end{tikzpicture}
					}}
					\\
					&G6 &0.52&0.62&0.71&\textbf{0.98}        &0.66&0.69&0.86&\textbf{0.98}  \\
					% G-9
					\multirow{2}{*}{\scalebox{0.5}{
							\begin{tikzpicture}
								\node[draw, circle, fill=cyan] (A) at (0,0){};
								\node[draw, circle, fill=cyan] (B) at (1,0){};
								\node[draw, circle, fill=cyan] (C) at (0,-1){};
								\node[draw, circle, fill=cyan] (D) at (1,-1){};
								\draw[->, line width=2pt, color=cyan] (A) -- (D); 
								\draw[->, line width=2pt, color=cyan] (A) -- (C); 
								\draw[->, line width=2pt, color=cyan] (C) -- (D);
							\end{tikzpicture}
					}}
					\\
					&G7 &0.39&0.49&0.71&\textbf{0.75}        &0.70&0.73&0.89&\textbf{0.89}  \\
					% G-10
					\multirow{2}{*}{\scalebox{0.5}{
							\begin{tikzpicture}
								\node[draw, circle, fill=brown] (A) at (0,0){};
								\node[draw, circle, fill=brown] (B) at (1,0){};
								\node[draw, circle, fill=brown] (C) at (0,-1){};
								\node[draw, circle, fill=brown] (D) at (1,-1){};
								\draw[->, line width=2pt, color=brown] (A) -- (B); 
								\draw[->, line width=2pt, color=brown] (A) -- (C); 
								\draw[->, line width=2pt, color=brown] (B) -- (D);
								\draw[->, line width=2pt, color=brown] (C) -- (D);
							\end{tikzpicture}
					}}
					\\
					&G8 &0.47&0.54&0.50&\textbf{0.59}          &0.6&0.63&0.62&\textbf{0.68} \\
					% G-11
					\multirow{2}{*}{\scalebox{0.5}{
							\begin{tikzpicture}
								\node[draw, circle, fill=gray] (A) at (0,0){};
								\node[draw, circle, fill=gray] (B) at (1,0){};
								\node[draw, circle, fill=gray] (C) at (0,-1){};
								\node[draw, circle, fill=gray] (D) at (1,-1){};
								\draw[->, line width=2pt, color=gray] (A) -- (D); 
								\draw[->, line width=2pt, color=gray] (B) -- (D); 
								\draw[->, line width=2pt, color=gray] (C) -- (D);
							\end{tikzpicture}
					}}
					\\
					&G9 &0.30&0.68&0.83&\textbf{0.85}          &0.40&0.76&0.86&\textbf{0.87}  \\
					% G-10
					\multirow{2}{*}{\scalebox{0.5}{
							\begin{tikzpicture}
								\node[draw, circle, fill=magenta] (A) at (0,0){};
								\node[draw, circle, fill=magenta] (B) at (1,0){};
								\node[draw, circle, fill=magenta] (C) at (0,-1){};
								\node[draw, circle, fill=magenta] (D) at (1,-1){};
								\draw[->, line width=2pt, color=magenta] (A) -- (B); 
								\draw[->, line width=2pt, color=magenta] (B) -- (C); 
								\draw[->, line width=2pt, color=magenta] (A) -- (D);
							\end{tikzpicture}
					}}
					\\
					&G10 &0.39&0.39&\textbf{0.52} &{0.32}         &0.53&0.56&\textbf{0.82}&{0.70}  \\
					\bottomrule
				\end{tabular}
			\end{threeparttable}
		\end{small}
	\end{adjustbox}
\caption{Comparison of identifiability results}
\label{tab:D4}
\end{table}
%\end{minipage}%
The results in Table~\ref{tab:D4} indicate that our method ICRL-SM can identify the true causal graph in most cases. The worst results are seen for graphs $G5$ and $G10$. As mentioned in \cite{causal_survey, sparsity_MIS}, causal graphs are sparse and in the $G5$ case, where the graph is fully connected, the proposed method cannot identify the causal variables well. Furthermore, in the next experiment we are going to examine the factors affecting causal disentanglement such as the number of edges in the graph and the intensity of soft intervention effect. These findings can explain why ICRL-SM cannot identify causal variables in $G10$ despite its sparsity.

\begin{table*}[!h]
	\centering
	\caption{Table comparing action and object accuracy across various methods on Causal-Triplet datasets under different settings. $z$ and $z_i$ show whether all causal variables ($z$), or only the intervened casual variable ($z_i$) are used for the prediction task. $R_{64}$ denote images with resolutions $64 \times 64$. }
	\label{tab:real_accuracy}
	%\resizebox{10cm}{!}{
		%\begin{small}
			\begin{tabular}{l cccc cccc}
				\toprule
				& \multicolumn{4}{c}{\textbf{Epic-Kitchens}} & \multicolumn{4}{c}{\textbf{ProcTHOR}} \\
				\cmidrule(lr){2-5} \cmidrule(lr){6-9}
				
				&\multicolumn{2}{c}{\textbf{Action Accuracy}} & 
				\multicolumn{2}{c}{\textbf{Object Accuracy}} & 
				\multicolumn{2}{c}{\textbf{Action Accuracy}} & 
				\multicolumn{2}{c}{\textbf{Object Accuracy}} \\
				\cmidrule(lr){2-3} \cmidrule(lr){4-5}
				\cmidrule(lr){6-7} \cmidrule(lr){8-9}
				\textbf{Method} & 
				\multicolumn{1}{c}{\textbf{$z$;$R_{64}$}} &
				\multicolumn{1}{c}{\textbf{$z_i$;$R_{64}$}} &
				\multicolumn{1}{c}{\textbf{$z$;$R_{64}$}} &
				\multicolumn{1}{c}{\textbf{$z_i$;$R_{64}$}} &
				\multicolumn{1}{c}{\textbf{$z$;$R_{64}$}} &
				\multicolumn{1}{c}{\textbf{$z_i$;$R_{64}$}} &
				\multicolumn{1}{c}{\textbf{$z$;$R_{64}$}} &
				\multicolumn{1}{c}{\textbf{$z_i$;$R_{64}$}} \\
				
				\midrule
				{$\beta$-VAE} \cite{beta-VAE}
				&\textbf{0.27} &0.18      &0.19&0.06   &\textbf{0.39}&0.30   &\textbf{0.44}&0.37 \\
    
				{dVAE} \cite{dvae}   
				&0.19&0.69      &\textbf{0.20}&0.17   &0.35&0.81  &0.40&0.78     \\
				
				ILCM \cite{ILCM}
				&0.21&0.59      &0.14&0.14   &0.30&0.70   &0.41&0.76     \\   
    
				\textbf{ICRL-SM (ours)}   
				&0.16&\textbf{0.86}      &0.16&\textbf{0.18}   &0.28&\textbf{0.93}   &0.40&\textbf{0.82}   \\
				
				%\addlinespace   
				\bottomrule
			\end{tabular}
		%\end{small}
	%}
\end{table*}

\subsection{Factors Affecting Causal Disentanglement}
In this experiment, we consider the graph $G3$, which has the best identifiability, and change the intensity of soft intervention and number of edges in its data generation process. To change the intensity, the post-intervention $\widetilde{loc}$ network weights are initialized with samples drawn from $N(1,1)$ (almost similar to $loc$) and $N(10,1)$ (significantly different from $loc$). To change the number of edges, we consider a chain and fully-connected graph.

\begin{table}[!h]
    %\begin{minipage}{\textwidth}
      \caption{Action and object accuracy of the explicit models with experiments conducted applying an image with resolution of $R_{64}$ as the input to the Resnet50 encoder with the intervened causal variable ($z_i$). Fixed-order denotes that the causal variables' order in the adjacency matrix remains unchanged.}
        \label{tab:explicit_softcd}
        \centering
        %\begin{minipage}{0.45\textwidth}
            \centering
            %\resizebox{\textwidth}{!}{%
                \begin{tabular}{ll cc}
                    \toprule
                    \textbf{Datasets} & \textbf{Methods} & \textbf{Action Accuracy} & \textbf{Object Accuracy} \\
                    \midrule
                    Epic-Kitchens
                    & ENCO \cite{enco}       &0.69      &0.13      \\
                    & DDS \cite{dds}         &0.44      &0.09         \\
                    & Fixed-order        &0.79      &0.14         \\
                    & \textbf{ICRL-SM (ours)}   &\textbf{0.86}      &\textbf{0.18} \\
                    \midrule
                    ProcTHOR
                    & ENCO \cite{enco}       &0.45       &0.53   \\
                    & DDS \cite{dds}         &0.64       &0.67    \\
                    & Fixed-order        &0.65      &0.54         \\
                    & \textbf{ICRL-SM (ours)}   &\textbf{0.93}       &\textbf{0.82}\\
                    \bottomrule
                \end{tabular}%
            %}
        %\end{minipage}%
 \end{table}
 \hfill % Add horizontal space between the minipages
 %\hspace{2em}
\begin{table}[!h] 
        \caption*{Action and object accuracy}
        %\begin{minipage}{0.45\textwidth}
            \centering
            %\begin{small}
                %\resizebox{\textwidth}{!}{%
                    \begin{tabular}{c c c c}
                        \toprule
                        \textbf{Edges} & \textbf{Post-intervention} & \textbf{Causal} & \textbf{Causal} \\
                        & \textbf{causal mechanism} & \textbf{Disentanglement} & \textbf{Completeness} \\
                        \midrule
                        Chain & Default & 0.98 & 0.98 \\
                        Full & Default & 0.89 & 0.89 \\
                        Default & Significantly different & 0.68 & 0.73 \\
                        Default & Almost similar & 0.85 & 0.86 \\  
                        \bottomrule
                    \end{tabular}%
                %}
                \caption*{ICRL-SM performance on different configurations of $G5$}
                \label{tab:ablation}
            %\end{small}
        %\end{minipage}
    %\end{minipage}
    \caption{Table shows the comparison of ICRL-SM performance on different configurations of $G5$ in Table~\ref{tab:D4}.}
    \label{tab:comparison}
\end{table}

The results in Table \ref{tab:comparison} further confirms the sparsity of causal graphs as the causal disentanglement is much worse in the fully-connected graph than the default graph of $G3$. The result for significantly different post-intervention causal mechanisms indicate that the switch variable cannot approximate intense effects of soft intervention and more supervision is required to observe $V$. Similar post-intervention causal mechanisms also do not have sufficient variability to disentangle the causal variables as mentioned in Theory \ref{theory:ident}.  

\subsection{Action Inference}
In this experiment, we show the performance of ICRL-SM in the real-world Causal-Triplet datasets. In these datasets $V$ i.e., soft intervention effects, are not directly observable. Nevertheless, our findings suggest that incorporating soft interventions through $V$ leads to superior performance compared to other implicit modeling approaches. Clearly, understanding the impact of soft interventions on the generative system of the dataset would result in improved outcomes.

The results in Table~\ref{tab:real_accuracy} indicate that when including all causal variables to predict actions, ICRL-SM performs at par with the baseline methods. However, including all causal variables in the action or object inference may cause spurious correlations. Therefore, we have also experimented with including only the related causal variable in action and object inference. In this setting, ICRL-SM significantly outperforms the baseline methods which means that it can better disentangle the causal variables. We have also compared ICRL-SM with explicit causal representation learning methods. ENCO \cite{enco} and DDS \cite{dds} have variable topological order of causal variables during training. Furthermore, we have included a specific setting where the topological order is fixed during training.  As shown in Table \ref{tab:comparison}, our proposed method has superior performance to explicit models as well.

\subsection{Scalability}
\begin{figure}[!h]
	\centering    
	\includegraphics[width=0.6\textwidth]{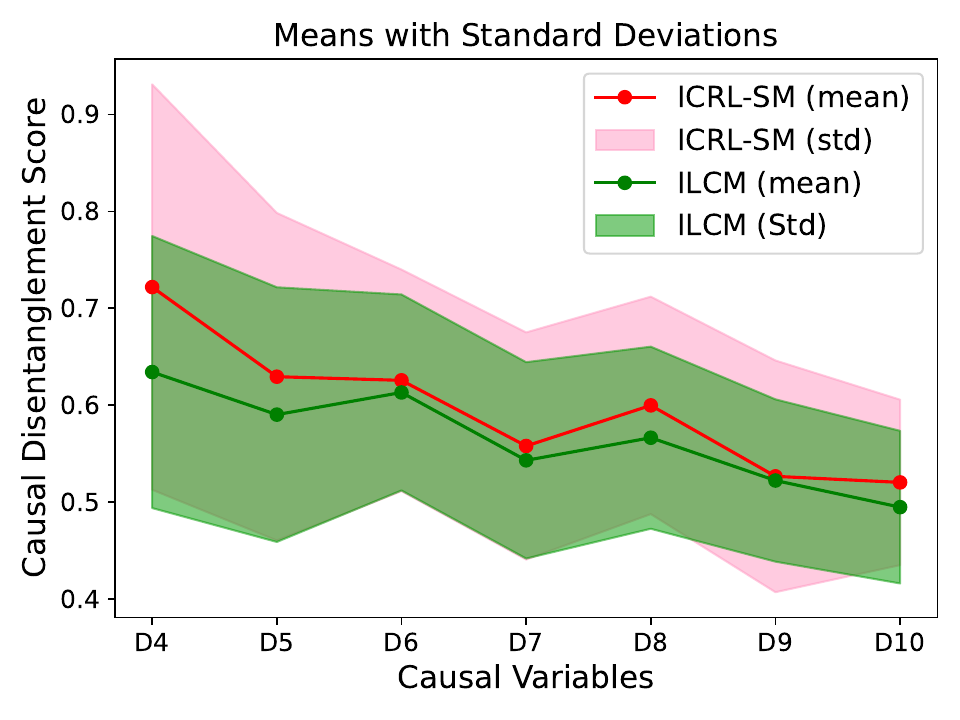}
	\caption{Causal disentanglement for different number of causal variables.}
	\label{fig:mean-std}
\end{figure}

While our primary research objective centered on addressing identifiability challenges in implicit causal models under soft interventions, we also conducted an investigation into the scalability of our proposed model. To comprehensively assess its performance, we designed experiments covering a range of causal graphs, featuring 5 to 10 variables, with 10 different seeds for each variable, following a similar experimental setup as our 4-variable causal graph experiments. The outcomes of these experiments, comparing ICRL-SM and ILCM, are presented in Figure~\ref{fig:mean-std}. By increasing the number of variables in the graph, confounding factors and ambiguities of causal relations increase as well. Consequently, more supervision on $\mathcal{V}$ is required to better separate the effect of causal variables themselves on the observed variables.

\section{Conclusion}
ICRL-SM, our novel model, enhances implicit causal representation learning during soft interventions by introducing a causal mechanism switch variable. Evaluations on synthetic and real-world datasets demonstrate ICRL-SM's superiority over state-of-the-art methods, highlighting its practical effectiveness. Our findings emphasize ICRL-SM's ability to discern causal models from soft interventions, marking it as a promising avenue for future research.

%{\small
%	\bibliographystyle{plain}
%	\setlength{\bibsep}{0pt}
%	\bibliography{references}
%}
%
%%%%%%%%%%%%%%%%%%%%%%%%%%%%%%%%%%%%%%%%%%%%%%%%%%%%%%%%%%%%%%%%%%%%%%%%%%%%%
%% APPENDIX
%%%%%%%%%%%%%%%%%%%%%%%%%%%%%%%%%%%%%%%%%%%%%%%%%%%%%%%%%%%%%%%%%%%%%%%%%%%%%
%\clearpage
%\input{Appendix}
%
%%%%%%%%%%%%%%%%%%%%%%%%%%%%%%%%%%%%%%%%%%%%%%%%%%%%%%%%%%%%%
%% CHECK LIST
%%%%%%%%%%%%%%%%%%%%%%%%%%%%%%%%%%%%%%%%%%%%%%%%%%%%%%%%%%%%%
%\clearpage
%\input{neurips_check_list}

\clearpage
\bibliographystyle{unsrtnat}
\bibliography{references}  

\begin{thebibliography}{40}
\providecommand{\natexlab}[1]{#1}
\providecommand{\url}[1]{\texttt{#1}}
\expandafter\ifx\csname urlstyle\endcsname\relax
  \providecommand{\doi}[1]{doi: #1}\else
  \providecommand{\doi}{doi: \begingroup \urlstyle{rm}\Url}\fi

\bibitem[Sch{\"{o}}lkopf(2019)]{causality_ML}
Bernhard Sch{\"{o}}lkopf.
\newblock Causality for machine learning.
\newblock \emph{CoRR}, abs/1911.10500, 2019.

\bibitem[Yang et~al.(2021{\natexlab{a}})Yang, Liu, Chen, Shen, Hao, and
  Wang]{causalVAE}
Mengyue Yang, Furui Liu, Zhitang Chen, Xinwei Shen, Jianye Hao, and Jun Wang.
\newblock Causalvae: Disentangled representation learning via neural structural
  causal models.
\newblock In \emph{{IEEE} Conference on Computer Vision and Pattern
  Recognition, {CVPR}}, pages 9593--9602. Computer Vision Foundation / {IEEE},
  2021{\natexlab{a}}.
\newblock \doi{10.1109/CVPR46437.2021.00947}.

\bibitem[Ahuja et~al.(2023)Ahuja, Mahajan, Wang, and
  Bengio]{interventional_survey}
Kartik Ahuja, Divyat Mahajan, Yixin Wang, and Yoshua Bengio.
\newblock Interventional causal representation learning.
\newblock In \emph{International Conference on Machine Learning, {ICML}},
  volume 202 of \emph{Proceedings of Machine Learning Research}, pages
  372--407. {PMLR}, 2023.

\bibitem[Brehmer et~al.(2022)Brehmer, de~Haan, Lippe, and Cohen]{ILCM}
Johann Brehmer, Pim de~Haan, Phillip Lippe, and Taco~S. Cohen.
\newblock Weakly supervised causal representation learning.
\newblock In \emph{NeurIPS}, 2022.

\bibitem[von K\"{u}gelgen et~al.(2021)von K\"{u}gelgen, Sharma, Gresele,
  Brendel, Sch\"{o}lkopf, Besserve, and Locatello]{causal3dident}
Julius von K\"{u}gelgen, Yash Sharma, Luigi Gresele, Wieland Brendel, Bernhard
  Sch\"{o}lkopf, Michel Besserve, and Francesco Locatello.
\newblock Self-supervised learning with data augmentations provably isolates
  content from style.
\newblock In M.~Ranzato, A.~Beygelzimer, Y.~Dauphin, P.S. Liang, and J.~Wortman
  Vaughan, editors, \emph{Advances in Neural Information Processing Systems},
  volume~34, pages 16451--16467. Curran Associates, Inc., 2021.
\newblock URL
  \url{https://proceedings.neurips.cc/paper_files/paper/2021/file/8929c70f8d710e412d38da624b21c3c8-Paper.pdf}.

\bibitem[Lippe et~al.(2022{\natexlab{a}})Lippe, Magliacane, L{\"{o}}we, Asano,
  Cohen, and Gavves]{citris}
Phillip Lippe, Sara Magliacane, Sindy L{\"{o}}we, Yuki~M. Asano, Taco Cohen,
  and Stratis Gavves.
\newblock {CITRIS:} causal identifiability from temporal intervened sequences.
\newblock In \emph{International Conference on Machine Learning, {ICML}},
  volume 162 of \emph{Proceedings of Machine Learning Research}, pages
  13557--13603. {PMLR}, 2022{\natexlab{a}}.

\bibitem[Pearl et~al.(2016)Pearl, Glymour, and P.~Jewell]{causaljudea}
Judea Pearl, Madelyn Glymour, and Nicholas P.~Jewell.
\newblock \emph{Causal inference in statistics: A primer}.
\newblock John Wiley and Sons, 2016.

\bibitem[Correa and Bareinboim(2020)]{soft_transportability}
Juan~D. Correa and Elias Bareinboim.
\newblock General transportability of soft interventions: Completeness results.
\newblock In Hugo Larochelle, Marc'Aurelio Ranzato, Raia Hadsell,
  Maria{-}Florina Balcan, and Hsuan{-}Tien Lin, editors, \emph{Advances in
  Neural Information Processing Systems 33: Annual Conference on Neural
  Information Processing Systems, NeurIPS}, 2020.

\bibitem[Bagi et~al.(2023)Bagi, Gharaee, Schulte, and Crowley]{GCRL}
Shayan Shirahmad~Gale Bagi, Zahra Gharaee, Oliver Schulte, and Mark Crowley.
\newblock Generative causal representation learning for out-of-distribution
  motion forecasting.
\newblock In \emph{International Conference on Machine Learning, {ICML}},
  volume 202 of \emph{Proceedings of Machine Learning Research}, pages
  31596--31612. {PMLR}, 2023.

\bibitem[Liu et~al.(2021)Liu, Sun, Wang, Tang, Li, Qin, Chen, and Liu]{csg}
Chang Liu, Xinwei Sun, Jindong Wang, Haoyue Tang, Tao Li, Tao Qin, Wei Chen,
  and Tie-Yan Liu.
\newblock Learning causal semantic representation for out-of-distribution
  prediction.
\newblock In M.~Ranzato, A.~Beygelzimer, Y.~Dauphin, P.S. Liang, and J.~Wortman
  Vaughan, editors, \emph{Advances in Neural Information Processing Systems},
  volume~34, pages 6155--6170. Curran Associates, Inc., 2021.

\bibitem[Liu et~al.(2022)Liu, Cadei, Schweizer, Bahmani, and
  Alahi]{causalmotion}
Yuejiang Liu, Riccardo Cadei, Jonas Schweizer, Sherwin Bahmani, and Alexandre
  Alahi.
\newblock Towards robust and adaptive motion forecasting: {A} causal
  representation perspective.
\newblock In \emph{{IEEE/CVF} Conference on Computer Vision and Pattern
  Recognition, {CVPR} 2022, New Orleans, LA, USA, June 18-24, 2022}, pages
  17060--17071. {IEEE}, 2022.
\newblock \doi{10.1109/CVPR52688.2022.01657}.
\newblock URL \url{https://doi.org/10.1109/CVPR52688.2022.01657}.

\bibitem[Yang et~al.(2021{\natexlab{b}})Yang, Yu, Cao, Liu, Wang, and Li]{CAE}
Shuai Yang, Kui Yu, Fuyuan Cao, Lin Liu, Hao Wang, and Jiuyong Li.
\newblock Learning causal representations for robust domain adaptation.
\newblock \emph{IEEE Transactions on Knowledge and Data Engineering}, pages
  1--1, 2021{\natexlab{b}}.
\newblock \doi{10.1109/TKDE.2021.3119185}.

\bibitem[Yu et~al.(2019)Yu, Chen, Gao, and Yu]{DAG-GNN}
Yue Yu, Jie Chen, Tian Gao, and Mo~Yu.
\newblock {DAG}-{GNN}: {DAG} structure learning with graph neural networks.
\newblock In \emph{Proceedings of the 36th International Conference on Machine
  Learning}, volume~97 of \emph{Proceedings of Machine Learning Research},
  pages 7154--7163. PMLR, 2019.

\bibitem[Lachapelle et~al.(2022)Lachapelle, Rodr{\'{\i}}guez, Sharma, Everett,
  Priol, Lacoste, and Lacoste{-}Julien]{sparse_disentanglement}
S{\'{e}}bastien Lachapelle, Pau Rodr{\'{\i}}guez, Yash Sharma, Katie Everett,
  R{\'{e}}mi~Le Priol, Alexandre Lacoste, and Simon Lacoste{-}Julien.
\newblock Disentanglement via mechanism sparsity regularization: {A} new
  principle for nonlinear {ICA}.
\newblock In \emph{1st Conference on Causal Learning and Reasoning, {CLeaR}},
  volume 177 of \emph{Proceedings of Machine Learning Research}, pages
  428--484. {PMLR}, 2022.

\bibitem[Schölkopf et~al.(2021)Schölkopf, Locatello, Bauer, Ke, Kalchbrenner,
  Goyal, and Bengio]{causal_survey}
Bernhard Schölkopf, Francesco Locatello, Stefan Bauer, Nan~Rosemary Ke, Nal
  Kalchbrenner, Anirudh Goyal, and Yoshua Bengio.
\newblock Toward causal representation learning.
\newblock \emph{Proceedings of the IEEE}, 109\penalty0 (5):\penalty0 612--634,
  2021.
\newblock \doi{10.1109/JPROC.2021.3058954}.

\bibitem[Kaddour et~al.(2022)Kaddour, Lynch, Liu, Kusner, and
  Silva]{crl_survey}
Jean Kaddour, Aengus Lynch, Qi~Liu, Matt~J. Kusner, and Ricardo Silva.
\newblock Causal machine learning: {A} survey and open problems.
\newblock \emph{CoRR}, abs/2206.15475, 2022.
\newblock \doi{10.48550/arXiv.2206.15475}.

\bibitem[Lu et~al.(2022)Lu, Wu, Hern{\'{a}}ndez{-}Lobato, and
  Sch{\"{o}}lkopf]{icarl}
Chaochao Lu, Yuhuai Wu, Jos{\'{e}}~Miguel Hern{\'{a}}ndez{-}Lobato, and
  Bernhard Sch{\"{o}}lkopf.
\newblock Invariant causal representation learning for out-of-distribution
  generalization.
\newblock In \emph{The Tenth International Conference on Learning
  Representations, {ICLR}}, 2022.

\bibitem[Shen et~al.(2022)Shen, Liu, Dong, Lian, Chen, and Zhang]{DEAR}
Xinwei Shen, Furui Liu, Hanze Dong, Qing Lian, Zhitang Chen, and Tong Zhang.
\newblock Weakly supervised disentangled generative causal representation
  learning.
\newblock \emph{J. Mach. Learn. Res.}, 23:\penalty0 241:1--241:55, 2022.

\bibitem[Yu et~al.(2020)Yu, Guo, Liu, Li, Wang, Ling, and Wu]{markovblanket}
Kui Yu, Xianjie Guo, Lin Liu, Jiuyong Li, Hao Wang, Zhaolong Ling, and Xindong
  Wu.
\newblock Causality-based feature selection: Methods and evaluations.
\newblock \emph{ACM Comput. Surv.}, 53\penalty0 (5), 2020.
\newblock ISSN 0360-0300.
\newblock \doi{10.1145/3409382}.

\bibitem[Perry et~al.(2022)Perry, von K{\"{u}}gelgen, and
  Sch{\"{o}}lkopf]{sparsity_MIS}
Ronan Perry, Julius von K{\"{u}}gelgen, and Bernhard Sch{\"{o}}lkopf.
\newblock Causal discovery in heterogeneous environments under the sparse
  mechanism shift hypothesis.
\newblock In \emph{NeurIPS}, 2022.

\bibitem[Zheng et~al.(2018)Zheng, Aragam, Ravikumar, and
  Xing]{acyclic_constraint}
Xun Zheng, Bryon Aragam, Pradeep Ravikumar, and Eric~P. Xing.
\newblock Dags with {NO} {TEARS:} continuous optimization for structure
  learning.
\newblock In \emph{Advances in Neural Information Processing Systems 31: Annual
  Conference on Neural Information Processing Systems {NeurIPS}}, pages
  9492--9503, 2018.

\bibitem[Jaber et~al.(2020)Jaber, Kocaoglu, Shanmugam, and
  Bareinboim]{maximal_ancestral_Graphs}
Amin Jaber, Murat Kocaoglu, Karthikeyan Shanmugam, and Elias Bareinboim.
\newblock Causal discovery from soft interventions with unknown targets:
  Characterization and learning.
\newblock In \emph{Advances in Neural Information Processing Systems 33: Annual
  Conference on Neural Information Processing Systems, NeurIPS}, 2020.

\bibitem[Cooper and Yoo(2013)]{cooper2013causal}
Gregory~F. Cooper and Changwon Yoo.
\newblock Causal discovery from a mixture of experimental and observational
  data, 2013.

\bibitem[Zhang et~al.(2023)Zhang, Greenewald, Squires, Srivastava, Shanmugam,
  and Uhler]{causaldiscrepancy}
Jiaqi Zhang, Kristjan~H. Greenewald, Chandler Squires, Akash Srivastava,
  Karthikeyan Shanmugam, and Caroline Uhler.
\newblock Identifiability guarantees for causal disentanglement from soft
  interventions.
\newblock In \emph{Advances in Neural Information Processing Systems 36: Annual
  Conference on Neural Information Processing Systems 2023, NeurIPS 2023, New
  Orleans, LA, USA, December 10 - 16, 2023}, 2023.
\newblock URL
  \url{http://papers.nips.cc/paper\_files/paper/2023/hash/9d3a4cdf6f70559e8c6fe02170fba568-Abstract-Conference.html}.

\bibitem[Wendong et~al.(2023)Wendong, Kekić, von Kügelgen, Buchholz,
  Besserve, Gresele, and Schölkopf]{cauca}
Liang Wendong, Armin Kekić, Julius von Kügelgen, Simon Buchholz, Michel
  Besserve, Luigi Gresele, and Bernhard Schölkopf.
\newblock Causal component analysis, 2023.

\bibitem[Squires et~al.(2023)Squires, Seigal, Bhate, and
  Uhler]{intervention_linear_cd}
Chandler Squires, Anna Seigal, Salil Bhate, and Caroline Uhler.
\newblock Linear causal disentanglement via interventions, 2023.

\bibitem[Varici et~al.(2023)Varici, Acarturk, Shanmugam, Kumar, and
  Tajer]{interventions_score_based}
Burak Varici, Emre Acarturk, Karthikeyan Shanmugam, Abhishek Kumar, and Ali
  Tajer.
\newblock Score-based causal representation learning with interventions, 2023.

\bibitem[Locatello et~al.(2020)Locatello, Poole, R{\"{a}}tsch, Sch{\"{o}}lkopf,
  Bachem, and Tschannen]{dvae}
Francesco Locatello, Ben Poole, Gunnar R{\"{a}}tsch, Bernhard Sch{\"{o}}lkopf,
  Olivier Bachem, and Michael Tschannen.
\newblock Weakly-supervised disentanglement without compromises.
\newblock In \emph{Proceedings of the 37th International Conference on Machine
  Learning,{ICML}}, volume 119 of \emph{Proceedings of Machine Learning
  Research}, pages 6348--6359. {PMLR}, 2020.

\bibitem[Lippe et~al.(2022{\natexlab{b}})Lippe, Cohen, and Gavves]{enco}
Phillip Lippe, Taco Cohen, and Efstratios Gavves.
\newblock Efficient neural causal discovery without acyclicity constraints.
\newblock In \emph{The Tenth International Conference on Learning
  Representations, {ICLR}}. OpenReview.net, 2022{\natexlab{b}}.

\bibitem[Charpentier et~al.(2022)Charpentier, Kibler, and G{\"{u}}nnemann]{dds}
Bertrand Charpentier, Simon Kibler, and Stephan G{\"{u}}nnemann.
\newblock Differentiable {DAG} sampling.
\newblock In \emph{The Tenth International Conference on Learning
  Representations, {ICLR}}. OpenReview.net, 2022.

\bibitem[Higgins et~al.(2017)Higgins, Matthey, Pal, Burgess, Glorot, Botvinick,
  Mohamed, and Lerchner]{beta-VAE}
Irina Higgins, Lo{\"{\i}}c Matthey, Arka Pal, Christopher~P. Burgess, Xavier
  Glorot, Matthew~M. Botvinick, Shakir Mohamed, and Alexander Lerchner.
\newblock beta-vae: Learning basic visual concepts with a constrained
  variational framework.
\newblock In \emph{5th International Conference on Learning Representations,
  {ICLR}}, 2017.

\bibitem[Buchholz et~al.(2023)Buchholz, Rajendran, Rosenfeld, Aragam,
  Schölkopf, and Ravikumar]{intervention_linear_nonlinearmixing}
Simon Buchholz, Goutham Rajendran, Elan Rosenfeld, Bryon Aragam, Bernhard
  Schölkopf, and Pradeep Ravikumar.
\newblock Learning linear causal representations from interventions under
  general nonlinear mixing, 2023.

\bibitem[von Kügelgen et~al.(2023)von Kügelgen, Besserve, Wendong, Gresele,
  Kekić, Bareinboim, Blei, and Schölkopf]{unknownI_nonparamteric}
Julius von Kügelgen, Michel Besserve, Liang Wendong, Luigi Gresele, Armin
  Kekić, Elias Bareinboim, David~M. Blei, and Bernhard Schölkopf.
\newblock Nonparametric identifiability of causal representations from unknown
  interventions, 2023.

\bibitem[Zheng et~al.(2022)Zheng, Ng, and Zhang]{nonlinearICA}
Yujia Zheng, Ignavier Ng, and Kun Zhang.
\newblock On the identifiability of nonlinear {ICA:} sparsity and beyond.
\newblock In \emph{NeurIPS}, 2022.

\bibitem[Pearl(2005)]{causality}
Judea Pearl.
\newblock \emph{Causality}, cambridge university press {(2000)}.
\newblock \emph{Artif. Intell.}, 169\penalty0 (2):\penalty0 174--179, 2005.

\bibitem[Immer et~al.(2023)Immer, Schultheiss, Vogt, Sch{\"{o}}lkopf,
  B{\"{u}}hlmann, and Marx]{lsnm}
Alexander Immer, Christoph Schultheiss, Julia~E. Vogt, Bernhard
  Sch{\"{o}}lkopf, Peter B{\"{u}}hlmann, and Alexander Marx.
\newblock On the identifiability and estimation of causal location-scale noise
  models.
\newblock In \emph{International Conference on Machine Learning, {ICML}},
  volume 202 of \emph{Proceedings of Machine Learning Research}, pages
  14316--14332. {PMLR}, 2023.

\bibitem[Liu et~al.(2023)Liu, Alahi, Russell, Horn, Zietlow, Sch{\"{o}}lkopf,
  and Locatello]{causal_triplet}
Yuejiang Liu, Alexandre Alahi, Chris Russell, Max Horn, Dominik Zietlow,
  Bernhard Sch{\"{o}}lkopf, and Francesco Locatello.
\newblock Causal triplet: An open challenge for intervention-centric causal
  representation learning.
\newblock In \emph{Conference on Causal Learning and Reasoning, {CLeaR}},
  volume 213 of \emph{Proceedings of Machine Learning Research}, pages
  553--573. {PMLR}, 2023.

\bibitem[Deitke et~al.(2022)Deitke, VanderBilt, Herrasti, Weihs, Ehsani,
  Salvador, Han, Kolve, Kembhavi, and Mottaghi]{deitke2022}
Matt Deitke, Eli VanderBilt, Alvaro Herrasti, Luca Weihs, Kiana Ehsani, Jordi
  Salvador, Winson Han, Eric Kolve, Aniruddha Kembhavi, and Roozbeh Mottaghi.
\newblock Procthor: Large-scale embodied ai using procedural generation.
\newblock \emph{Advances in Neural Information Processing Systems},
  35:\penalty0 5982--5994, 2022.

\bibitem[Damen et~al.(2022)Damen, Doughty, Farinella, Furnari, Kazakos, Ma,
  Moltisanti, Munro, Perrett, Price, and Wray]{epcikitchens}
Dima Damen, Hazel Doughty, Giovanni~Maria Farinella, Antonino Furnari,
  Evangelos Kazakos, Jian Ma, Davide Moltisanti, Jonathan Munro, Toby Perrett,
  Will Price, and Michael Wray.
\newblock Rescaling egocentric vision: Collection, pipeline and challenges for
  {EPIC-KITCHENS-100}.
\newblock \emph{Int. J. Comput. Vis.}, 130\penalty0 (1):\penalty0 33--55, 2022.
\newblock \doi{10.1007/s11263-021-01531-2}.

\bibitem[Eastwood and Williams(2018)]{dci_scores}
Cian Eastwood and Christopher K.~I. Williams.
\newblock A framework for the quantitative evaluation of disentangled
  representations.
\newblock In \emph{6th International Conference on Learning Representations,
  {ICLR}}, 2018.

\end{thebibliography}

\clearpage
%%%%%%%%%%%%%%%%%%%%%%%%%%%%%%%%%%%%%%%%%%%%%%%%%%%%%%%%%%%%%%%%%%%%%%%%%%%%
% APPENDIX
%%%%%%%%%%%%%%%%%%%%%%%%%%%%%%%%%%%%%%%%%%%%%%%%%%%%%%%%%%%%%%%%%%%%%%%%%%%%

\renewcommand{\appendixpagename}{Appendix}
\begin{appendices}
	\renewcommand{\thesection}{A\arabic{section}} 
	\setcounter{table}{0}
	\renewcommand{\thetable}{A\arabic{table}}
	\setcounter{figure}{0}
	\renewcommand{\thefigure}{A\arabic{figure}}

	\section{Proof of Identifiability Theorem}
	\label{append:proof}

     \subsection{Notation Conventions}
        For better and easier understanding of our proposed methodology, including the underlying theory and its proof, we adhere to the notation conventions presented in Table~\ref{tab:notation} throughout the paper:
        \begin{table}[h!]
	\centering
	\caption{Notation Conventions}
	
	\begin{tabular}{ll}
		\toprule
		\textbf{Symbol} & \textbf{Description} \\
		\midrule
		{Lowercase letters} ($x$) & Values of variables \\
		{Uppercase letters} ($X$) & Variables themselves \\
		{Calligraphic letters} ($\mathcal{X}$) & Domains of variables and sets \\
		$i$ & General index for variables \\
		$i \in I$ & Index for intervention targets among causal variables \\
		$\tilde{(.)}$ & Post-intervention state \\
		$({.})_{/i}$ & Exclusion of the $i$-th element \\
		\bottomrule
	\end{tabular}
 \label{tab:notation}
\end{table}

In order to prove our model is identifiable we need a two additional definitions and some previously stated assumptions.

    \begin{definition} \textbf{Structural Causal Models}\\
        \label{def:scm}
        A structural causal model (SCM) is a tuple $\mathcal{C}=(\mathcal{F}, \mathcal{Z}, \mathcal{E}, \mathcal{G})$ with the following components:
        \begin{enumerate}[noitemsep]
         \item The domain of causal variables $\mathcal{Z}=\mathcal{Z}_1 \times \mathcal{Z}_2 \times \ldots \times \mathcal{Z}_n$.
         \item The domain of exogenous variables $\mathcal{E}=\mathcal{E}_1 \times \mathcal{E}_2 \times \ldots \times \mathcal{E}_n$.
         \item A directed acyclic graph $\mathcal{G}(\mathcal{C})$ over the causal and exogenous variables.
         \item A causal mechanism $f_i \in \mathcal{F}$ which maps an assignment of parent values for the parents $Z_{pa_i}$ plus an exogenous variable value for $E_i$ to a value of causal variable $Z_i$. 
        \end{enumerate}
    
    \end{definition}

    \begin{definition}(Component-wise Transformation)
        \label{def:component-wise}
        Let $\phi$  be a transformation (1-1 onto mapping) between product spaces:
        \begin{equation}
        \phi: \Pi_{i=1}^n \mathcal{X}_i \rightarrow \Pi_{i=1}^n \mathcal{Y}_i.
        \end{equation}
        If there exist local transformations %components 
        %$\phi_i \in \phi$ 
        $\phi_i$ such that:
        \begin{equation}
        \forall i,j, \: \forall x, \: \phi(x_1, x_2, ..., x_n)_i = \phi_i(x_j),
        \end{equation}
        then $\phi$ is a component-wise transformation.
    \end{definition}
    
    \begin{definition}(Diffeomorphism)
         A diffeomorphism between smooth manifolds $\mathcal{M}$ and $\mathcal{N}$ is a bijective map $f: \mathcal{M} \rightarrow \mathcal{N}$, which is smooth and has a smooth inverse. Diffeomorphisms preserve information as they are invertible transformations without discontinuous changes in their image. 
    \end{definition}

    \begin{definition}(Pushforward measure)
    \label{def:pushmeasure}
        Given a measurable function $f: \mathcal{A} \rightarrow \mathcal{B}$ between two measurable spaces $\mathcal{A}$ and $\mathcal{B}$, and a measure $p$ defined on $\mathcal{A}$, the pushforward measure $f_*p$ on $\mathcal{B}$ is defined for measurable sets $E$ in $\mathcal{B}$ as:
        \begin{equation}
            (f_* p)(E)=p(f^{-1}(E)),
        \end{equation}
        where $_*$ denotes the pushforward operation. In other words, the pushforward measure $f_*p$ assigns a measure to a set in $\mathcal{B}$ by measuring the pre-image of that set under $f$ in the space $\mathcal{A}$.
    \end{definition}

    \begin{lemma}
    \label{lem:step2}
    The transformation $\phi_{\mathcal{Z}}: \mathcal{Z} \rightarrow \mathcal{Z'}$ between the causal variable of two LCMs $\mathcal{M}$ and $\mathcal{M'}$ defined in Definition \ref{def:equivalence} is a component-wise transformation, if $\: \forall i,j, i\neq j \quad \tilde{E'_i} \perp\!\!\!\perp \tilde{E'_j}$ and the causal variables follow a multivariate normal distribution conditional on the pre-intervention exogenous variables where $\tilde{E'_i}$ denote the post-intervention exogenous variable of causal variable $i$ in $\mathcal{M'}$. 

        proof:
        We consider the case where the exogenous variables are mapped to causal variables by a location-scale noise model such that:
       \begin{align*}
           \tilde{z}_i = \tilde{s}_i(\tilde{e}_i;e_{/i})= \frac{\tilde{e}_i - \widetilde{loc}(e_{/i})}{\widetilde{scale}(e_{/i})}, \quad \quad
         \forall i,j, i\neq j \quad \tilde{E'_i} \perp\!\!\!\perp \tilde{E'_j} \rightarrow \mathbb{E}\left[\tilde{E'_i}\tilde{E'_j}\right]=\mathbb{E}\left[\tilde{E'_i}\right]\mathbb{E}\left[\tilde{E'_j}\right],
        \end{align*}
        where $\mathbb{E}\left[.\right]$ denotes the expected value. 
        Let's add these three constants:
        \begin{equation}
                 - \mathbb{E}\left[\tilde{E'_i}\right]\widetilde{loc'_j}(e'_{/j}), \quad \: - \mathbb{E}\left[\tilde{E'_j}\right]\widetilde{loc'_i}(e'_{/i}), \quad \: \widetilde{loc'_i}(e'_{/i})\widetilde{loc'_j}(e'_{/j})
        \end{equation}
        to the both sides of the equality and then divide both sides by $\widetilde{scale'_i}(e'_{/i})\widetilde{scale'_j}(e'_{/j})$:
         \begin{align*}
            &\mathbb{E}\left[\frac{\tilde{E'_i}\tilde{E'_j} - \tilde{E'_i}\widetilde{loc'_j}(e'_{/j}) - \tilde{E'_j}\widetilde{loc'_i}(e'_{/i}) + \widetilde{loc'_i}(e'_{/i})\widetilde{loc'_j}(e'_{/j})}{\widetilde{scale'_i}(e'_{/i})\widetilde{scale'_j}(e'_{/j})}\right]= \\
            &\frac{\mathbb{E}\left[\tilde{E'_i}\right]\mathbb{E}\left[\tilde{E'_j}\right] -  \mathbb{E}\left[\tilde{E'_i}\right]\widetilde{loc'_j}(e'_{/j}) - \mathbb{E}\left[\tilde{E'_j}\right]\widetilde{loc'_i}(e'_{/i}) + \widetilde{loc'_i}(e'_{/i})\widetilde{loc'_j}(e'_{/j})}{\widetilde{scale'_i}(e'_{/i})\widetilde{scale'_j}(e'_{/j})} \\
            & \rightarrow \mathbb{E}\left[\left(\frac{\tilde{E'_i} - \widetilde{loc'_i}(e'_{/i})}{\widetilde{scale'_i}(e'_{/i})}\right) \left(\frac{\tilde{E'_j} - \widetilde{loc'_j}(e'_{/j})}{\widetilde{scale'_j}(e'_{/j})}\right) \right]= 
            \left(\frac{\mathbb{E}\left[\tilde{E'_i}\right] -\widetilde{loc'_i}(e'_{/i})}{\widetilde{scale'_i}(e'_{/i})}\right) \left(\frac{\mathbb{E}\left[\tilde{E'_j}\right] -\widetilde{loc'_j}(e'_{/j})}{\widetilde{scale'_j}(e'_{/j})}\right) \\
            &\rightarrow \mathbb{E}\left[\tilde{Z'_i}\tilde{Z'_j}|E'\right] = \mathbb{E}\left[\tilde{Z'_i}|E'\right]\mathbb{E}\left[\tilde{Z'_j}|E'\right] \\
            & \rightarrow \mathbb{E}\left[\tilde{Z'_i}\tilde{Z'_j}|E'\right] - \mathbb{E}\left[\tilde{Z'_i}|E\right]\mathbb{E}\left[\tilde{Z'_j}|E'\right] = 0 \\
            & \rightarrow \mathbb{E}\left[\tilde{Z'_i}\tilde{Z'_j}|E'\right] - \mathbb{E}\left[\tilde{Z'_i}|E'\right]\mathbb{E}\left[\tilde{Z'_j}|E'\right] - \mathbb{E}\left[\tilde{Z'_i}|E'\right]\mathbb{E}\left[\tilde{Z'_j}|E'\right] + \mathbb{E}\left[\tilde{Z'_i}|E'\right]\mathbb{E}\left[\tilde{Z'_j}|E'\right] = 0 \\
            & \rightarrow \mathbb{E}\left[\tilde{Z'_i}\tilde{Z'_j}|E'\right] - \mathbb{E}\left[\tilde{Z'_j}\mathbb{E}\left[\tilde{Z'_i}|E'\right]|E'\right] - \mathbb{E}\left[\tilde{Z'_i}\mathbb{E}\left[\tilde{Z'_j}|E'\right]|E'\right] + \mathbb{E}\left[\tilde{Z'_i}|E'\right]\mathbb{E}\left[\tilde{Z'_j}|E'\right] = 0 \\
            &\rightarrow \mathbb{E}\left[\left(\tilde{Z'_i} - \mathbb{E}\left[\tilde{Z'_i}|E'\right]\right)\left(\tilde{Z'_j} - \mathbb{E}\left[\tilde{Z'_j}|E'\right]\right)|E'\right]=0 \\
            & \rightarrow \text{cov}(\tilde{Z'_i}, \tilde{Z'_j}|E') = 0 
        \end{align*}

        Typically, the aforementioned equalities would be valid for any diffeomorphic solution function $\tilde{s}_i: \tilde{\mathcal{E}}_i \times \mathcal{E}_{/i} \rightarrow \tilde{\mathcal{Z}}_i$. However, in this paper, we specifically focus on solution functions represented by a location-scale noise model. 

        Assuming that the causal variables follow a \textbf{multivariate normal distribution conditional on the pre-intervention exogenous variables}, $\text{cov}(\tilde{Z'_i}, \tilde{Z'_j}|E') = 0$ would imply that $\Tilde{Z'_i} \perp\!\!\!\perp \Tilde{Z'_j} | E'$. Let's define $\phi_{\mathcal{E}}=g'^{-1} \circ g: \mathcal{E} \rightarrow \mathcal{E'}$ where $g$ and $g'$ are the decoders in $\mathcal{M}$ and $\mathcal{M'}$. As stated in Assumption \ref{assumption:ic}, the decoders are diffeomorphism, hence, $\phi_{\mathcal{E}}$ is a diffeomorphism. Furthermore, let's denote $\tilde{s}(e)=[\tilde{s}_1(e), \tilde{s}_2(e), ..., \tilde{s}_n(e)]$ as the set of all solution functions in post-intervention which are also diffeomorphism as stated in Assumption \ref{assumption:ic}. Consequently:

        \begin{align*}
            & (\text{$\phi_{\mathcal{E}}^{-1}$ is diffeomorphic}) \: \forall i,j, i\neq j \quad \Tilde{Z'_i} \perp\!\!\!\perp \Tilde{Z'_j} | E' \rightarrow \Tilde{Z'_i} \perp\!\!\!\perp \Tilde{Z'_j} | \phi_{\mathcal{E}}^{-1}(E') \rightarrow  \Tilde{Z'_i} \perp\!\!\!\perp \Tilde{Z'_j} | E \\
            &\rightarrow p(\Tilde{Z_i'} | E=e)p(\Tilde{Z'_j} | E=e) = p(\Tilde{Z_i'}, \Tilde{Z'_j} | E=e) \\
            & (\text{all functions in $\tilde{s}$ are diffeomorphism}) \rightarrow p(\Tilde{Z_i'} | \tilde{s}(e))p(\Tilde{Z'_j} | \tilde{s}(e)) = p(\Tilde{Z_i'}, \Tilde{Z'_j} | \tilde{s}(e)) \\
            &\rightarrow p(\Tilde{Z_i'} | \tilde{Z})p(\Tilde{Z'_j} | \tilde{Z}) = p(\Tilde{Z_i'}, \Tilde{Z'_j} | \tilde{Z})
        \end{align*}

        The association between $\tilde{\mathcal{Z}}'$ and $\tilde{\mathcal{Z}}$ arises from their shared observation space. We know that every causal variable in $\mathcal{M'}$ depends at least on one of the causal variables in $\mathcal{M}$. If one of the causal variables in $\mathcal{M'}$ depended on more than one causal variable in $\mathcal{M}$, it would create dependency between two variables in $\mathcal{M'}$ and violate the above equality. Therefore, no variable in $\mathcal{M'}$ depends on more than one causal variable in $\mathcal{M}$. Consequently, the transformation $\phi_\mathcal{Z}$ is a component-wise transformation.

    \end{lemma}

	\begin{theorem}(Identifiability of latent causal models.)
        \label{theory:ident_append}
        Let $\mathcal{M}=(\mathcal{A}, \mathcal{X}, g, \mathcal{I})$ and $\mathcal{M}'=(\mathcal{A}', \mathcal{X}, g', \mathcal{I})$ be two LCMs with shared observation space $\mathcal{X}$ and shared intervention targets $\mathcal{I}$. Suppose the following conditions are satisfied:
        \begin{enumerate}[noitemsep]
        \item Identical correspondence assumptions explained in \ref{assumption:ic}.
        \item Soft interventions satisfy  Assumption \ref{assumption:observev}.
        \item The causal and exogenous variables are real-valued. 
        \item The causal and exogenous variables follow a multivariate normal distribution. 
        \end{enumerate}
        Then the following statements are equivalent:
        \begin{enumerate}[noitemsep]
        \item Two LCMs $\mathcal{M}$ and $\mathcal{M}'$ assign the same likelihood to interventional and observational data i.e., $p^{\mathcal{X}}_{\mathcal{M}}(x,\Tilde{x}) = p^{\mathcal{X}}_{\mathcal{M}'}(x, \Tilde{x})$.\\ 
        \item $\mathcal{M}$ and $\mathcal{M}'$ are disentangled, that is $\mathcal{M} \sim_r \mathcal{M}'$ according to Definition \ref{def:equivalence}.
        \end{enumerate}
    \end{theorem}
	
	\mypara{Proof.}
	We will proceed to prove the equivalence between statements 1 and 2 by showing the implication is true in both forward and reverse directions.

    \subsection{Forward Direction: $\mathcal{M} \sim_r \mathcal{M'} \Rightarrow p^{\mathcal{X}}_{\mathcal{M}}(x,\Tilde{x}) = p^{\mathcal{X}}_{\mathcal{M'}}(x, \Tilde{x})$}
    This direction is fairly straightforward.   
    According to Definition \ref{def:equivalence}, the fact that $M \sim_r M'$ implies that $\phi_{\mathcal{E}}$ is measure preserving. Thus:
    
    \begin{equation}
        p^{\mathcal{E}'}_{\mathcal{M'}}(e', \Tilde{e}') = (\phi_{\mathcal{E}})_*p^{\mathcal{E}}_{\mathcal{M}}(e, \Tilde{e})
    \end{equation}

    Furthermore, considering that ancestry is preserved, $\phi_{\mathcal{Z}}$ is measure preserving, and that causal variables are obtained from their ancestral exogenous variables in implicit models, we have:
    
    \begin{equation}
        p^{\mathcal{Z}'}_{\mathcal{M'}}(z', \Tilde{z}') = (\phi_{\mathcal{Z}})_*p^{\mathcal{Z}}_{\mathcal{M}}(z, \Tilde{z})
    \end{equation}
    
    Since models are trained to maximize the log likelihood of $p(x, \Tilde{x}, \Tilde{x} - x)$ and the latent spaces in $M$ and $M'$ have the same distribution, the decoders should yield the same observational distributions as:
    
    \begin{equation}
        p^{\mathcal{X}}_{\mathcal{M}}(x,\Tilde{x}) = p^{\mathcal{X}}_{\mathcal{M'}}(x, \Tilde{x})
    \end{equation}
    
    \subsection{Reverse Direction: $p^{\mathcal{X}}_{\mathcal{M}}(x,\Tilde{x}) = p^{\mathcal{X}}_{\mathcal{M'}}(x, \Tilde{x}) \Rightarrow \mathcal{M} \sim_r \mathcal{M'}$}
    
    Let's define $\phi_{\mathcal{E}}=g'^{-1} \circ g: \mathcal{E} \rightarrow \mathcal{E'}$. Since we can express $e=s^{-1}(z)$, we can now define $\phi_{\mathcal{Z}}$ as: 
    \begin{align}
    \phi_{\mathcal{Z}}=s' \circ g'^{-1} \circ g \circ s^{-1}: \mathcal{Z} \rightarrow \mathcal{Z'}.
    \end{align}
    
    Therefore, we have: 
    \begin{equation}
        \phi_{\mathcal{E}} = s'^{-1} \circ \phi_{\mathcal{Z}} \circ s
    \end{equation}
    Because $g$ and $g'$ are \textbf{diffeomorphisms}, $\phi_{\mathcal{E}}$ is a diffeomorphism as well. Furthermore, since $p^{\mathcal{X}}_{\mathcal{M}}=p^{\mathcal{X}}_{\mathcal{M'}}$ and $\phi_{\mathcal{E}}$ is a diffeomorphism, then:
    
    \begin{equation}
        p^{\mathcal{E}'}_{\mathcal{M'}}=(\phi_{\mathcal{E}})_*p^{\mathcal{E}}_{\mathcal{M}}
    \end{equation}
    
    Consequently, $\phi_{\mathcal{E}}$ is measure-preserving. Similarly, $\phi_{\mathcal{Z}}$ is measure-preserving as well since causal mechanisms are \textbf{diffeomorphisms}.   

    \subsubsection{Step 1: Identical correspondence of edges and nodes}
    Let's define the set $U$ for \textbf{atomic} interventions $i \neq j \in I$ as:
    
    \begin{equation}
        U=\{(e, \tilde{e}) | supp \: p^{\mathcal{E}, \mathcal{I}}_{\mathcal{M}}(e,\Tilde{e}|i) \cap supp \: p^{\mathcal{E}, \mathcal{I}}_{\mathcal{M}}(e,\Tilde{e}|j)\} \subset \mathcal{E} \times \mathcal{E}
    \end{equation}
    
    Then, assuming \textbf{counterfactual exogenous variables}, we have:
    
    \begin{equation}
        p^{\mathcal{E}, \mathcal{I}}_{\mathcal{M}}(U|i)=p^{\mathcal{E}, \mathcal{I}}_{\mathcal{M}}(U|j)=0
    \end{equation}
    
    We also assume that \textbf{all interventions are observed}. Therefore, we can say that:
    
    \begin{equation}
        p_{\mathcal{M}}^{\mathcal{E}}(e, \Tilde{e})= \sum_{i \in I} p_{\mathcal{M}}^{\mathcal{E}, \mathcal{I}}(e, \Tilde{e}|i)p_{\mathcal{M}}^{\mathcal{I}}(i),
    \end{equation}
    
    is a discrete mixture of non-overlapping distributions $p_{\mathcal{M}}^{\mathcal{E}, \mathcal{I}}(e, \Tilde{e}|i)$. Similarly, we can say that $p_{\mathcal{M'}}^{\mathcal{E}'}(e', \Tilde{e}')$ is a discrete mixture of non-overlapping distributions. Since $\phi_{\mathcal{E}}$ is measure preserving, $p_{\mathcal{M}}^{\mathcal{E}}(e, \Tilde{e})$ = $p_{\mathcal{M'}}^{\mathcal{E}'}(e', \Tilde{e}')$. Therefore, 

    \begin{equation}
        \sum_{i \in I} p_{\mathcal{M}}^{\mathcal{E}, \mathcal{I}}(e, \Tilde{e}|i)p_{\mathcal{M}}^{\mathcal{I}}(i) = \sum_{i' \in I'} p_{\mathcal{M}'}^{\mathcal{E}', \mathcal{I}'}(e', \Tilde{e}'|i')p_{\mathcal{M}'}^{\mathcal{I}'}(i'),
    \end{equation}
    
    It can be concluded that as $\phi_{\mathcal{E}}$ must map between the conditional distributions, there exists a bijection that also induces a permutation $\psi: [n] \rightarrow [n]$.  %Note that if we had non-atomic interventions or non-counterfactual exogenous variables, then the conditional distributions would have some overlapping. With overlapping distributions, we can no longer claim there is a bijection mapping between the conditional distributions. 
    Additionally, since we assume that the \textbf{target of interventions are known} and we have a shared domain $\mathcal{I}$ in $\mathcal{M}$ and $\mathcal{M}'$, we have:

    \begin{equation}
        \sum_{i \in I} p_{\mathcal{M}}^{\mathcal{E}, \mathcal{I}}(e, \Tilde{e}|i)p_{\mathcal{M}}^{\mathcal{I}}(i) = \sum_{i' \in I'} p_{\mathcal{M}'}^{\mathcal{E}', \mathcal{I}}(e', \Tilde{e}'|i)p_{\mathcal{M}'}^{\mathcal{I}}(i),
    \end{equation}

    Consequently, the permutation $\psi$ is an identity transformation.
    
    In space $\mathcal{Z}$, the interventions should also be \textbf{sufficiently variable} in order to have non-overlapping  $p_{\mathcal{M}}^{\mathcal{Z}, \mathcal{I}}(z, \Tilde{z}|i)$ distributions. In the case of soft interventions, $\Tilde{z}$ is affected by all ancestral exogenous variables which could be ancestors of other causal variables as well. Consequently, if the changes in causal mechanisms are not sufficient, the effect of ancestral exogenous variables on causal variables will share some similarities and create overlapping distributions. Similar to space $\mathcal{E}$, we can say that there is a permutation between $p_{\mathcal{M}}^{\mathcal{Z}, \mathcal{I}}(z, \Tilde{z}|i)$ and $p_{\mathcal{M}'}^{\mathcal{Z}', \mathcal{I}'}(z', \Tilde{z}'|i')$ as well. Subsequently, as the target of interventions are known, the permutation is an identity transformation. 
    
\begin{figure}[!h]
	\centering
	\small
	\resizebox{\columnwidth}{!}{
	\begin{tabular}{cccc}
		\includegraphics[width=0.40\textwidth]{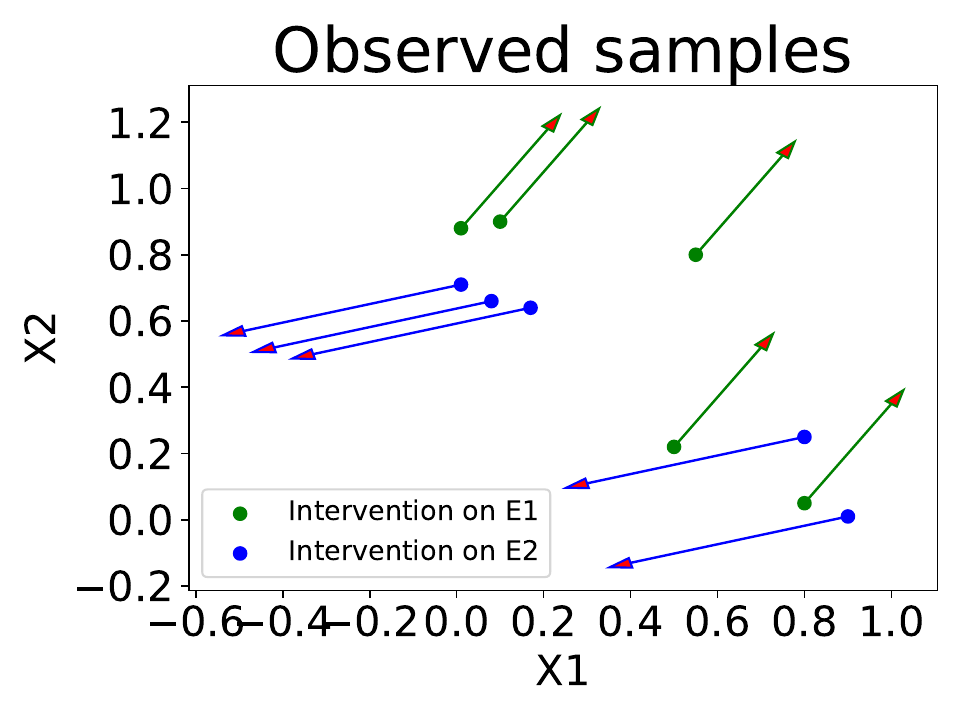}&
		\includegraphics[width=0.50\textwidth]{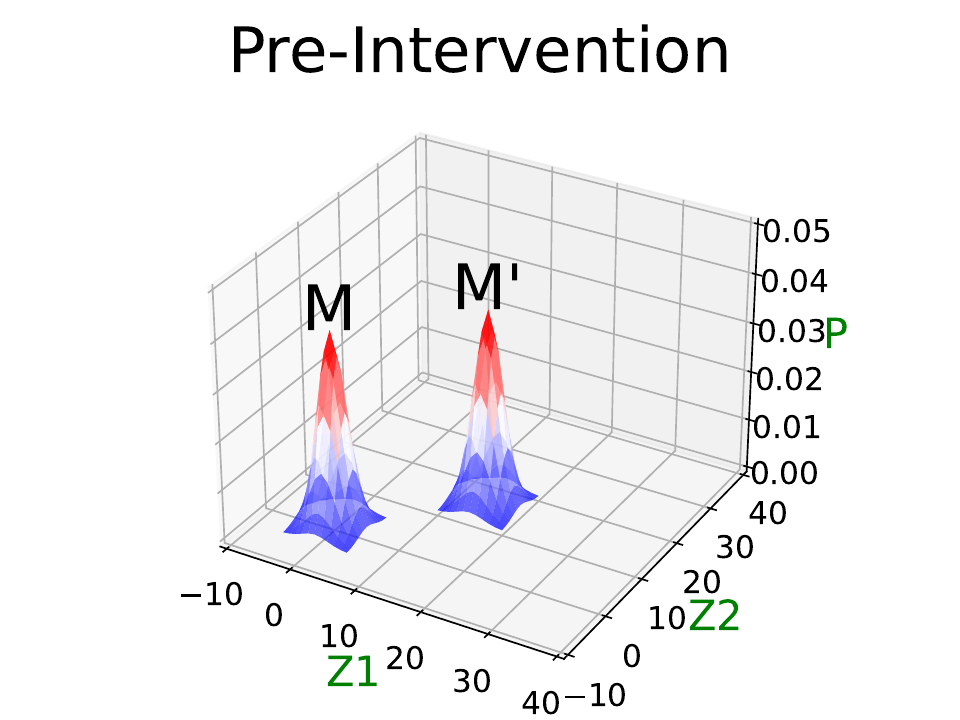} \\
          (a) & (b)\\
         \addlinespace[3mm]
		\includegraphics[width=0.50\textwidth]{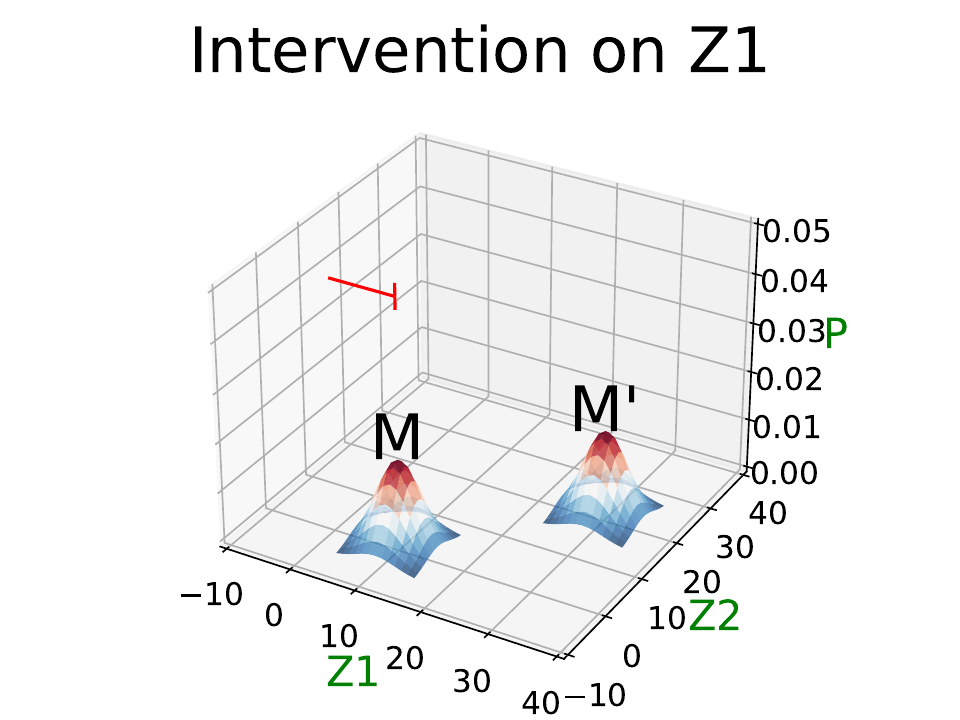} &
		\includegraphics[width=0.50\textwidth]{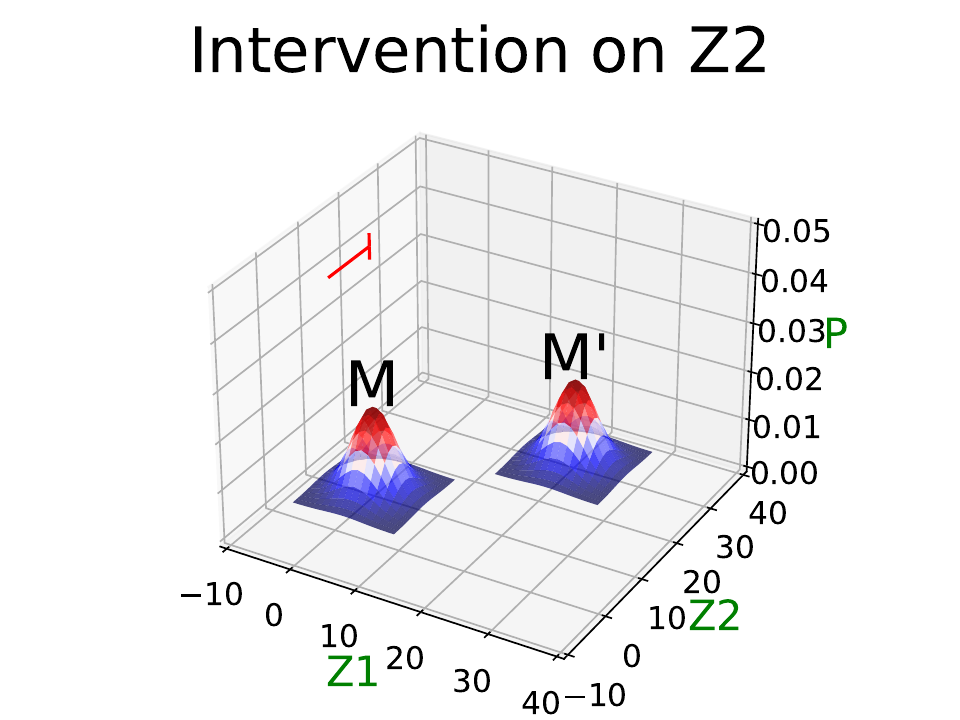} \\
        (c) & (d) \\
	\end{tabular}
}
	\caption{The distribution of observed and causal variables in two causal models $\mathcal{M}$ and $\mathcal{M'}$, which belong to the equivalence class up to reparameterization. (a) There are 10 observed samples in which $Z_1$ or $Z_2$ has been intervened on. (b) The distribution of causal variables when $I=0$ (no intervention) is identical to each other but the range of value of causal variables are different and can be mapped to each other using $\phi_{\mathcal{Z}}$. (c) The intervention on $Z_1$ ($I=1$). (d) The intervention on $Z_2$ ($I=2$). For $I=1$ and $I=2$ the distributions are again identical to each other but are different for different targets of intervention as soft interventions change the conditional distribution (condition on parents) of causal variables. Also, for each value of $I$, the distributions of $\mathcal{M}$ and $\mathcal{M'}$ should move in one direction as targets are known.}
	\label{fig:proof}
\end{figure}

    %When we are learning prior of causal variables, we can say that the joint distribution of causal variables is a mixture of non-overlapping distributions as explained above. By conditioning on $I$, we are telling the model to sample causal variables from one of these mixtures. Because the targets are known, we can write the equality above. So for example, in sample $(x_1, \Tilde{x}_1)$, we will be selecting the first conditional in both $\mathcal{M}$ and $\mathcal{M}'$ and not other conditionals.
    The effect of soft intervention with known targets on these conditional distributions is shown in Figure \ref{fig:proof}.

    \subsubsection{Step 2: Component-wise $\phi_{\mathcal{Z}}$}

According to Lemma \ref{lem:step2}, in order to prove that $\phi_{\mathcal{Z}}$ is a component-wise transformation, we need to prove that $\tilde{E'_i}$ and $\tilde{E'_j}$ are independent  $\forall i, j, i \neq j$. In implicit modeling we do not know the parents of each causal variable, hence, we assume the distribution of $\Tilde{Z'_i}$ to be conditioned only on $E'_i$ as in Equation \ref{eq:prior_tilde_e} since $E'_i$ is a known parent of $\Tilde{Z'_i}$. The mean of a conditional distribution can be calculated as:

\begin{equation}
    \label{eq:condition_mean}
\mathbb{E}[\Tilde{Z}'_i|\Tilde{E}'_i=e'_i]=\mu_{\Tilde{Z}'_i}+\rho \frac{\sigma_{\Tilde{Z}'_i}}{\sigma_{E'_i}}(e'_i - \mu_{E'_i})
\end{equation}

where $\rho$ and $\sigma$ are the correlation coefficient and variance of the random variables, respectively. On the other hand, we model $\Tilde{\mathcal{Z}'_i}$ using switch mechanisms as:

\begin{equation*}
    \Tilde{z}'_i = s_i(\Tilde{e}'_i; e'_{/i}, h(v'))
\end{equation*}

By using Taylor's expansion we can write above equation as:

\begin{align*}
&s_i(\Tilde{e}'_i; e'_{/i}, h_i(v')) = s_i(\Tilde{e}'_i; e'_{/i}, h_i(v'_0)) + + \sum_{n=1}^\infty \frac{1}{n!} \left( \frac{\partial^n s_i}{\partial h_i^n} \bigg|_{h_i=h_i(v'_0)} (h_i(v') - h_i(v'_0))^n \right) \\
&= s_i(\Tilde{e}'_i; e'_{/i}, h_i(v'_0)) + R_i
\end{align*}

Furthermore, we assume \textbf{separable dependence} such that:

\begin{equation*}
\exists v'_0 \text{ such that } \forall i \quad s_i(\tilde{e}'_i; e'_{/i}, h_i(v'_0)) = s_i(\tilde{e}'_i; e'_{/i})
\end{equation*}

An example of such a scenario could be in location-scale noise models, where a soft intervention changes the location parameter of the model as:

\begin{align*}
    &s_i(e'_i; e'_{/i})=e'_i + loc(e'_{/i}) \rightarrow \tilde{s_i}(\Tilde{e}'_i;e'_{/i}) =s_i(\Tilde{e}'_i;e'_{/i}, h_i(v')) \\
    &= \Tilde{e}'_i+loc(e'_{/i}) + h_i(v') = \Tilde{e}'_i + loc(e'_{/i}) + v'^2+v'
\end{align*}

In this example, for $v_0'=0$, $s_i(\tilde{e}'_i; e'_{/i}, h_i(v'_0)) = s_i(\tilde{e}'_i; e'_{/i})$. 

Consequently, we can write the following equality from Equation \ref{eq:condition_mean}:

\begin{align*}
    \mathbb{E}[\Tilde{Z'_i}|E'_i=e'_i]=\mathbb{E}[s'_i(\Tilde{E}'_i; E'_{/i}) + R_i|E'_i=e'_i] =\mu_{\Tilde{Z'_i}}+\rho \frac{\sigma_{\Tilde{Z'_i}}}{\sigma_{E'_i}}(e'_i - \mu_{E'_i}) 
\end{align*}

By taking the partial derivative of both side with respect to $\tilde{E'_j}$ we have:

\begin{align*}
   \forall j \neq i \quad \mathbb{E}[\frac{\partial s'_i(\tilde{E}'_i; E'_{/i})}{\partial \tilde{E}'_i} \cdot \frac{\partial \tilde{E}'_i}{\partial \tilde{E'_j}} + \frac{\partial s'_i(\tilde{E}'_i; E'_{/i})}{\partial E'_{/i}} \cdot \frac{\partial E'_{/i}}{\partial \tilde{E'_j}} + \frac{\partial R_i}{\partial \tilde{E'_j}} |E'_i=e'_i] = 0
\end{align*}

If we did not have the causal mechanism switch variable $h_i(v')$ (no $R_i$ term), the equation above would only hold if $s_i$ was constant in parents ($E_{/i}$), which is not the case due to the presence of soft interventions, or:
\begin{equation}
\frac{\partial s'_i(\tilde{E}'_i; E'_{/i})}{\partial \tilde{E}'_i} \cdot \frac{\partial \tilde{E}'_i}{\partial \tilde{E'_j}} =  - \frac{\partial s'_i(\tilde{E}'_i; E'_{/i})}{\partial E'_{/i}} \cdot \frac{\partial E'_{/i}}{\partial \tilde{E'_j}},
\end{equation}
then the latter scenario would imply that $\frac{\partial \Tilde{E'_i}}{\partial \tilde{E'_j}} \neq 0$, hence, $\tilde{E'_i} \not\perp\!\!\!\perp \tilde{E'_j}$. However, by introducing the causal mechanism switch variable $V$ and assuming it is observed, we can account for the effects of soft interventions through $h_i(v')$. In this case, $\frac{\partial \Tilde{E'_i}}{\partial \tilde{E'_j}} = 0$ as exogenous variables are commonly assumed to be independent in practice. Consequently:

\begin{align*}
    \forall i,j \quad \tilde{E'_i} \perp\!\!\!\perp \tilde{E'_j}  
    \quad \rightarrow \text{$\phi_{\mathcal{Z}}$ is a component-wise transformation.}
\end{align*}

    \textbf{Step 3: Component-wise $\phi_{\mathcal{E}}$} 

\begin{figure}[!h]
	\centering    
	\includegraphics[width=0.5\textwidth]{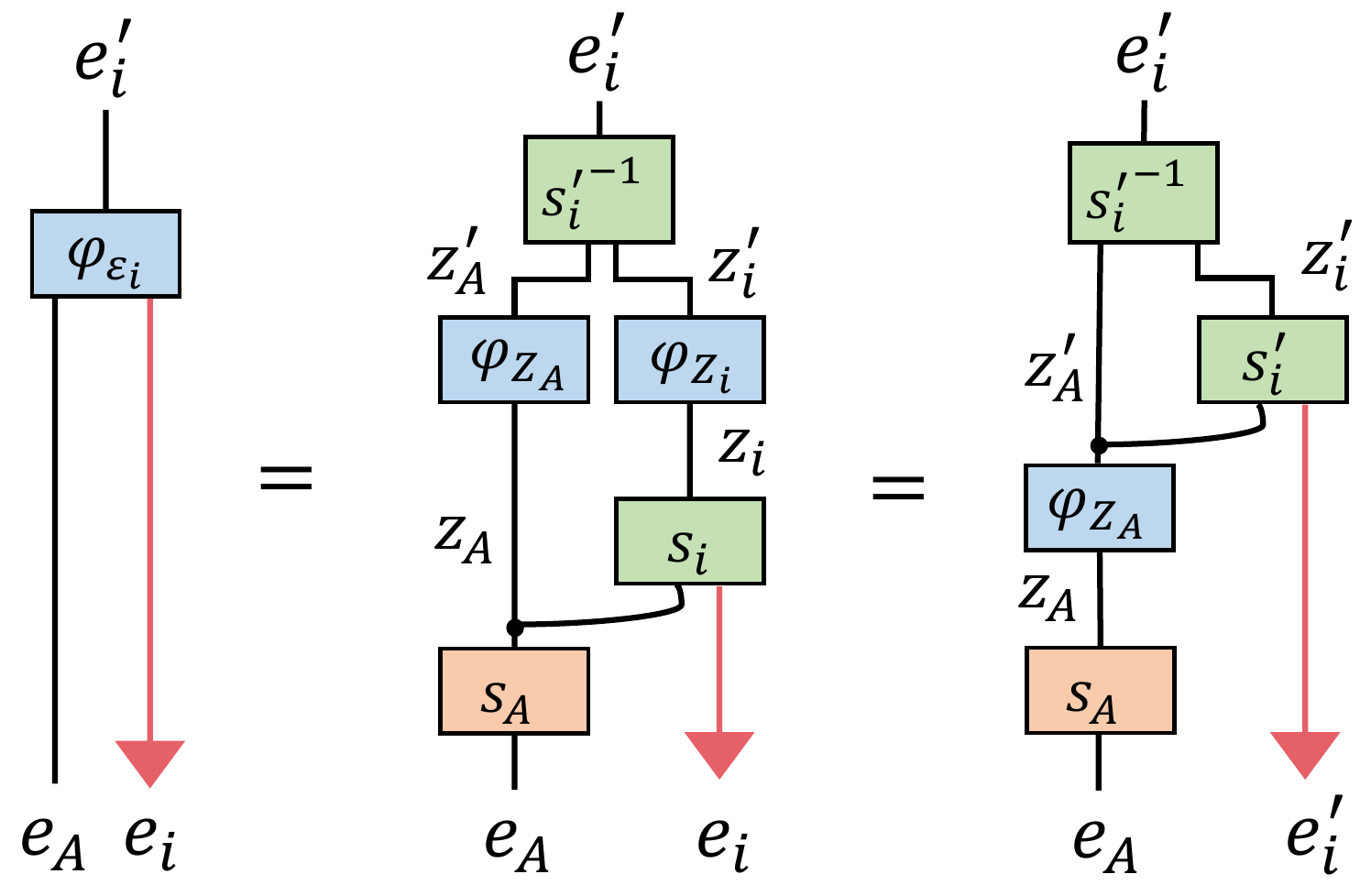}
	\caption{String diagrams for connections between $E$ and $E'$. The triangle indicates sampling variables from their corresponding distributions. }
	\label{fig:exogenous-connection}
\end{figure}

    Using the result from previous step that $\phi_{\mathcal{Z}}$ is a component-wise transformation, the string diagrams for connections between $E$ and $E'$ will be as shown in Figure \ref{fig:exogenous-connection}. $\phi_{\mathcal{E}}(e)_{i'} = (s'^{-1} \circ \phi_{\mathcal{Z}} \circ s)(e)_{i'}$ will only depend on $E_A$, where $A=anc_i$ is the ancestors of variable $i$, and $e_i$. Because:

    \begin{align*}
        & (\text{permutations of indices is identity}) \rightarrow \phi_{\mathcal{E}}(e)_{i'} = \phi_{\mathcal{E}}(e)_i \\
        & (\text{$s'^{-1}_i$ only depends on $z'_i$ and $z'_{anc_i}$}) \rightarrow \phi_{\mathcal{E}}(e)_i = s'^{-1}_i(z'_i, z'_{anc_i}) \\
        & (\text{$\phi_{\mathcal{Z}}$ is a component-wise transformation}) \rightarrow s'^{-1}_i(z'_i, z'_{anc_i}) = s'^{-1}_i(\phi_{\mathcal{Z}}(z_i)_i, \phi_{\mathcal{Z}}(z_{anc_i})_{anc_i}) \\
        & (\text{$s_i$ and $s_{anc_i}$ only depend on $e_i$ and $e_{anc_i}$}) \rightarrow s'^{-1}_i(\phi_{\mathcal{Z}}(z_i)_i, \phi_{\mathcal{Z}}(z_{anc_i})_{anc_i}) = \\
        & s'^{-1}_i(\phi_{\mathcal{Z}}(s_i(e_i, e_{anc_i}))_i, \phi_{\mathcal{Z}}(s_{anc_i}(e_{anc_i}))_{anc_i})
    \end{align*}
        
      Please note that the arguments of the solution functions presented above reflect their dependencies, rather than the precise inputs used in the implementation. The first equality in Figure \ref{fig:exogenous-connection} follows from the definition of $\phi_{\mathcal{E}_i}$. The second equality holds when we first apply $\phi_{\mathcal{Z}_A}$ and then apply the causal mechanisms. It can be concluded from the most right-hand side diagram in Figure \ref{fig:exogenous-connection} that the transformation from $\mathcal{E}'_i \times \mathcal{E}_A \rightarrow \mathcal{E}'_i$ is constant in $\mathcal{E}_A$. Therefore, $\phi_{\mathcal{E}_i}$ is a component-wise transformation.

%%%%%%%%%%%%%%%%%%%%%%%%%%%%%%%%
\begin{figure*}[!h]
\begin{center}
\small
%\begin{tabular}{m{.22\textwidth} m{.22\textwidth} m{.22\textwidth} m{.22\textwidth}}
\begin{tabularx}{1\textwidth} {
   >{\centering\arraybackslash}X 
   >{\centering\arraybackslash}X 
   >{\centering\arraybackslash}X 
   >{\centering\arraybackslash}X }
   
% %%%%%%%%%%%%%%%%%%%% Epic-Kitchens %%%%%%%%%%%%%
(a) Pre-Epic-Kitchens & (b) Pre-Epic-Kitchens & (c) Pre-Epic-Kitchens & (d) Pre-Epic-Kitchens\\
\includegraphics[width=0.20\textwidth]{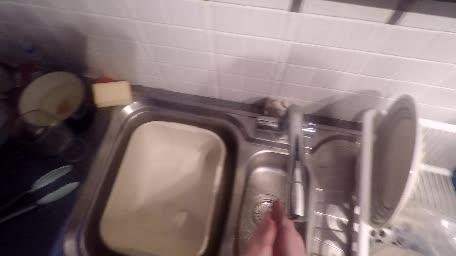} & \includegraphics[width=0.20\textwidth]{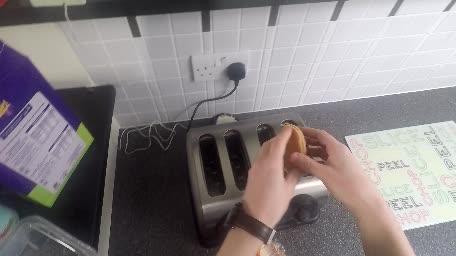} & \includegraphics[width=0.20\textwidth]{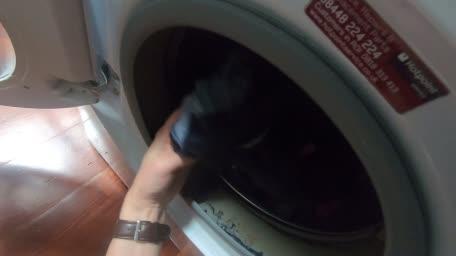} & \includegraphics[width=0.20\textwidth]{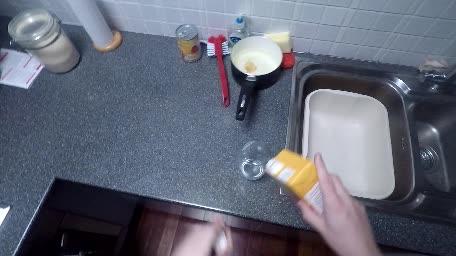}\\
%\textbf{896,324} & \textbf{89,311} & \textbf{47,328} & \textbf{46,970}\\
\addlinespace[1mm]
(a) Post:~Valve-locked & (b) Post:~Bread-Inserted & (c) Post:~Clothes-Gathered & (d) Post:~Juice-Poured\\
\includegraphics[width=0.20\textwidth]{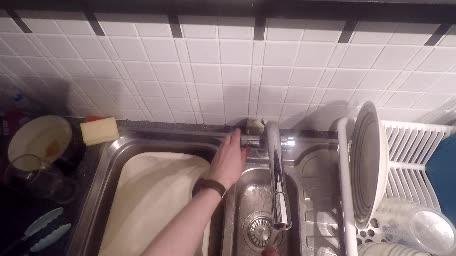} & \includegraphics[width=0.20\textwidth]{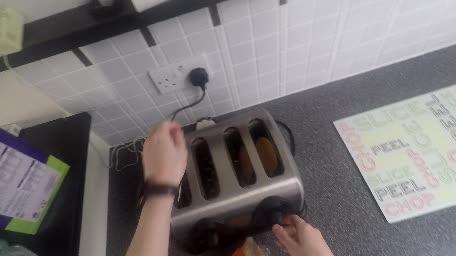} & \includegraphics[width=0.20\textwidth]{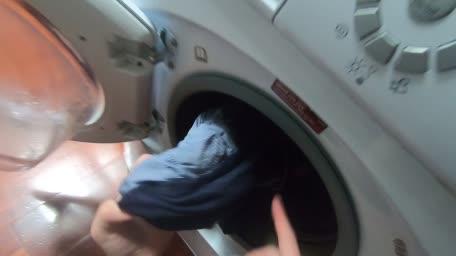} & \includegraphics[width=0.20\textwidth]{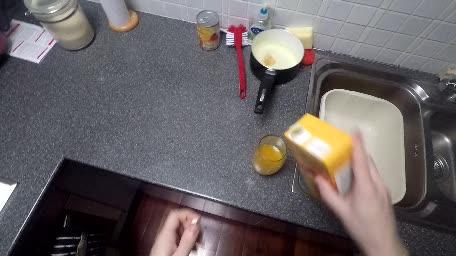}\\
\addlinespace[3mm]
% %%%%%%%%%%%%%%%%% ProcTHOR %%%%%%%%%%%%%%%%%%
(e) Pre-ProcTHOR & (f) Pre-ProcTHOR & (g) Pre-ProcTHOR & (h) Pre-ProcTHOR\\
\includegraphics[width=0.20\textwidth]{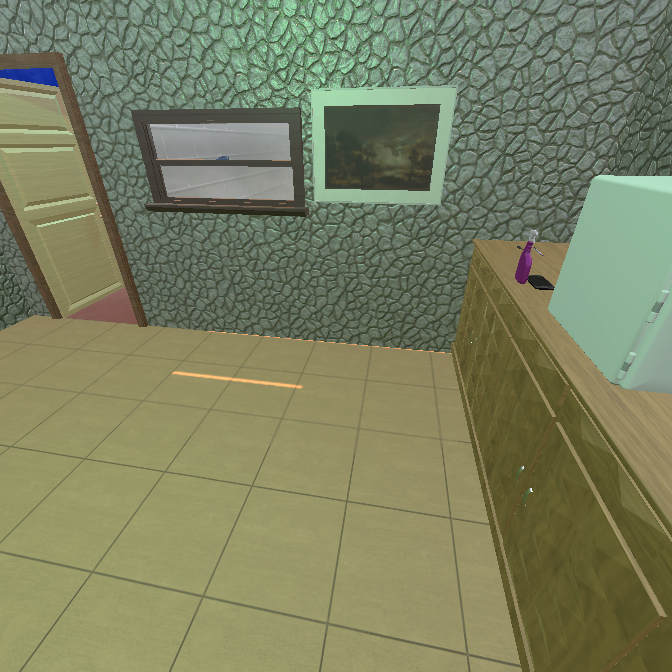} & \includegraphics[width=0.20\textwidth]{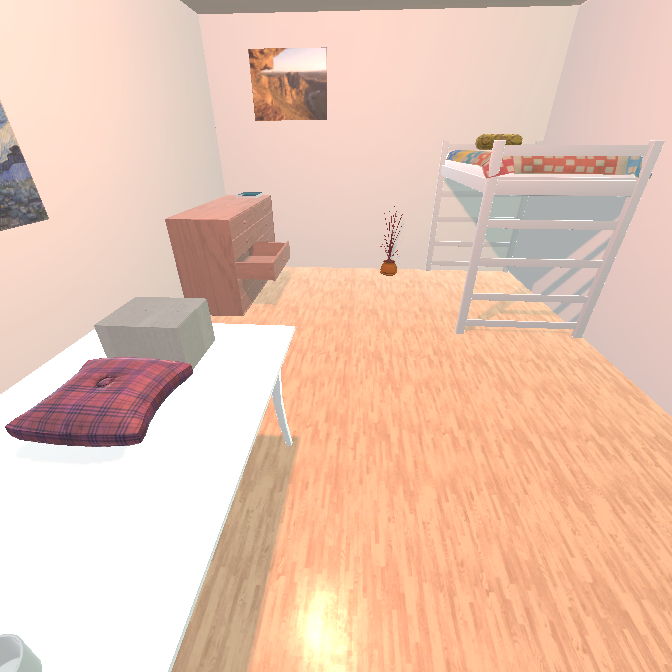} & \includegraphics[width=0.20\textwidth]{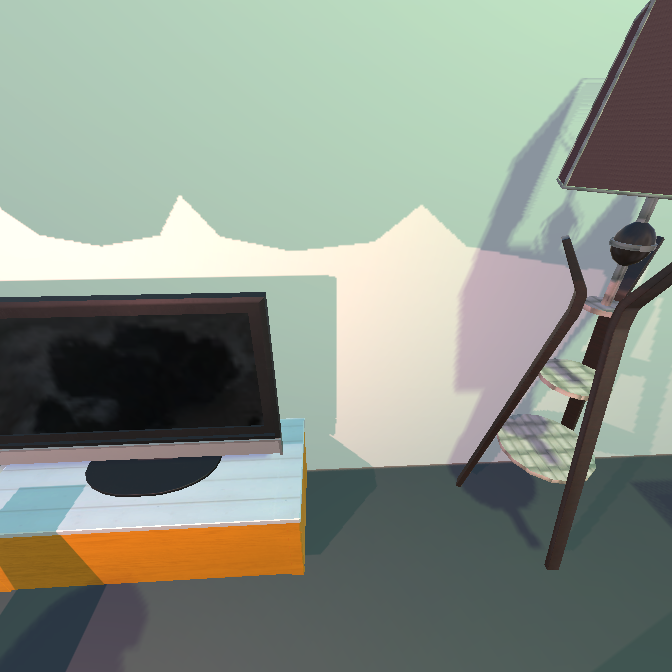} & \includegraphics[width=0.20\textwidth]{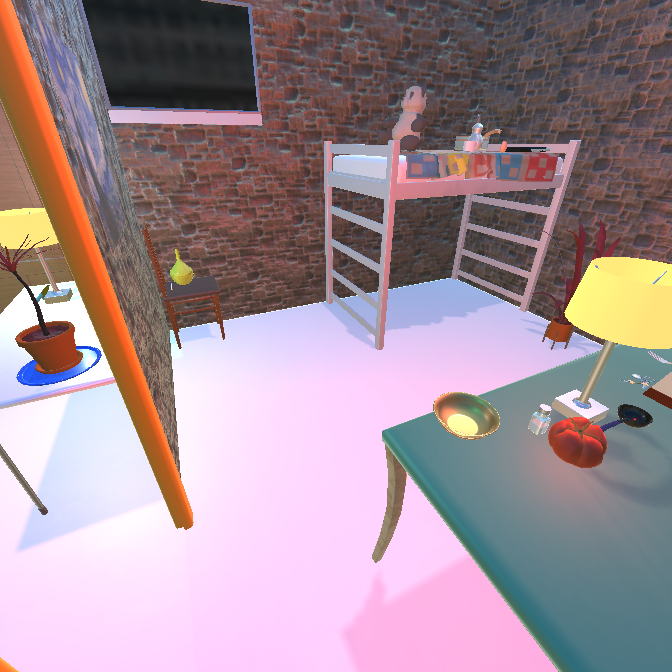}\\
\addlinespace[1mm]
(e) Post:~Cabinet-Open & (f) Post:~Box-Open & (g) Post:~TV-Broken & (h) Post:~TV-On\\
\includegraphics[width=0.20\textwidth]{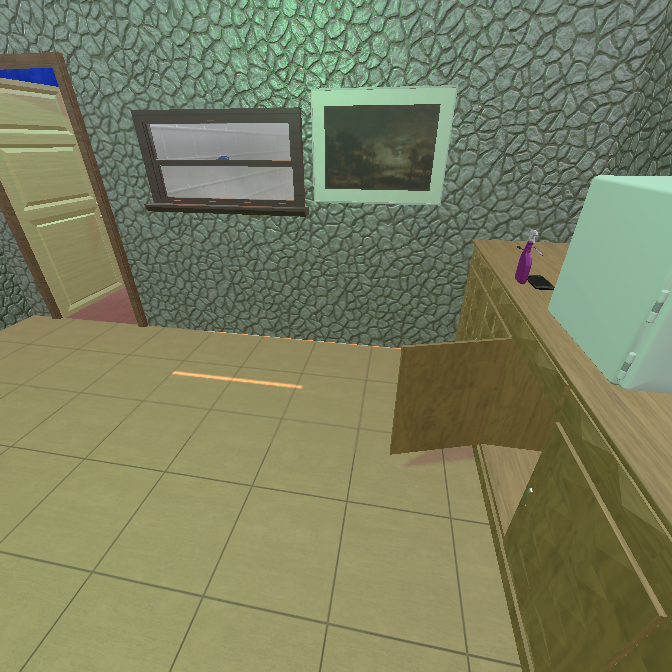} & \includegraphics[width=0.20\textwidth]{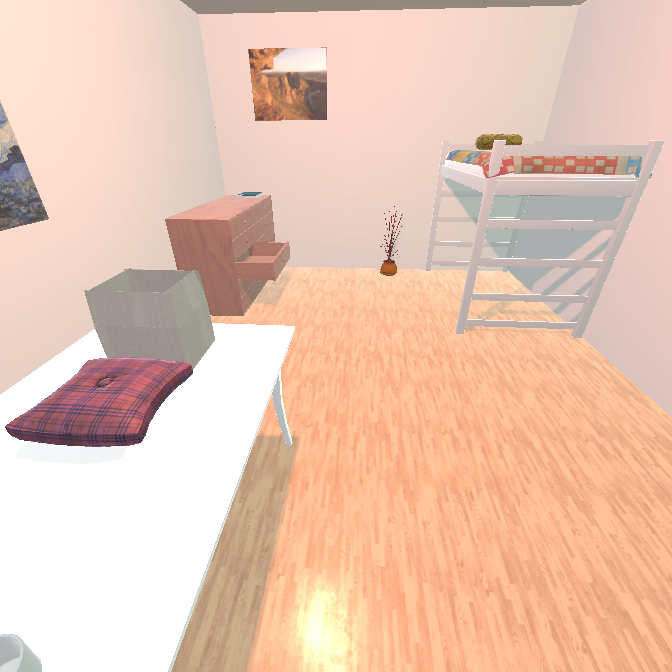} & \includegraphics[width=0.20\textwidth]{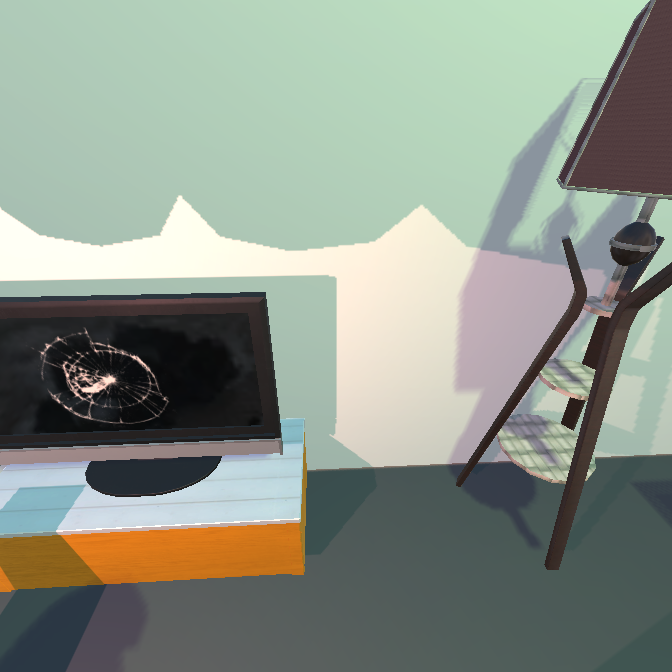} & \includegraphics[width=0.20\textwidth]{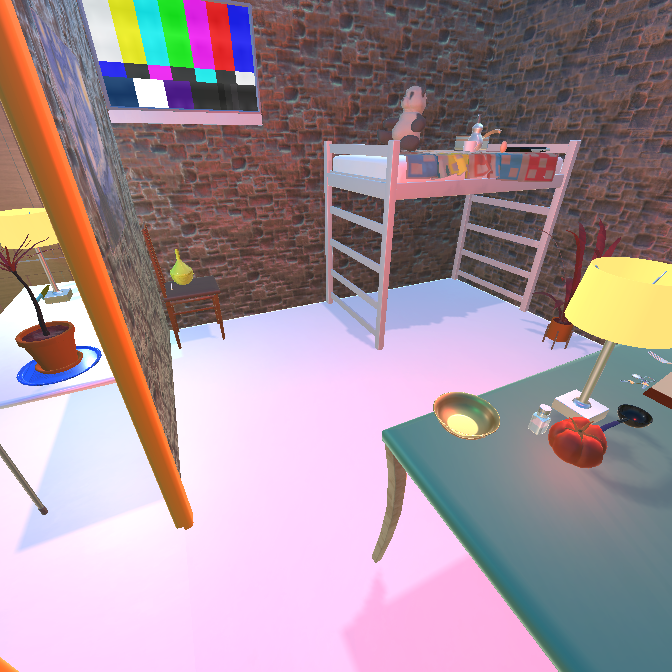}\\
\addlinespace[1mm]

\end{tabularx}
\end{center}
\caption{In the Causal-Triplet dataset \cite{causal_triplet}, visual representations capture both pre and post-intervention scenarios. The first two rows showcase data samples from Epic-Kitchens, while the third and fourth rows feature samples from ProcTHOR. Each image in the post-intervention condition is accompanied by labels specifying the corresponding action and intervened object. In the images in the first two rows, the agent is performing an action on an object but the camera angle has also changed. So we can say that for example the distribution of causal variables conditioned on the camera angle has been changed due to soft intervention.}
\label{fig:causal_triplet}
\end{figure*}

	\section{Soft vs. Hard intervention}
	
	In a causal model, an intervention refers to a deliberate action taken to manipulate or change one or more variables in order to observe its impact on other variables within the causal model. Interventions help to study how changes in one variable directly cause changes in another, thereby revealing causal relationships.

%\begin{wrapfigure}{r}{0.4\textwidth}
\begin{figure}
    \centering    
    \begin{subfigure}[b]{0.3\textwidth}
		\centering
		\includegraphics[width=0.55\textwidth]{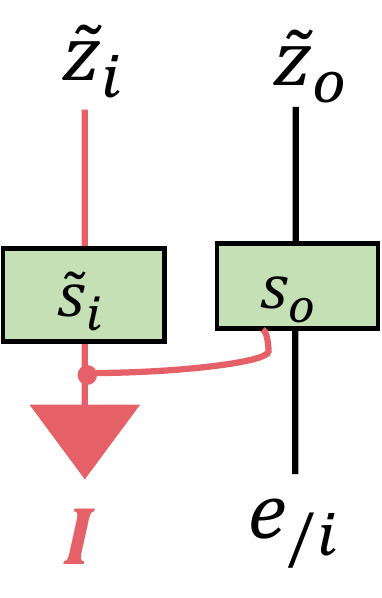}
		\caption{}
		\label{fig:hard}
	\end{subfigure}
	\hspace{1em}
	\begin{subfigure}[b]{0.3\textwidth}
		\centering
		\includegraphics[width=0.5\textwidth]{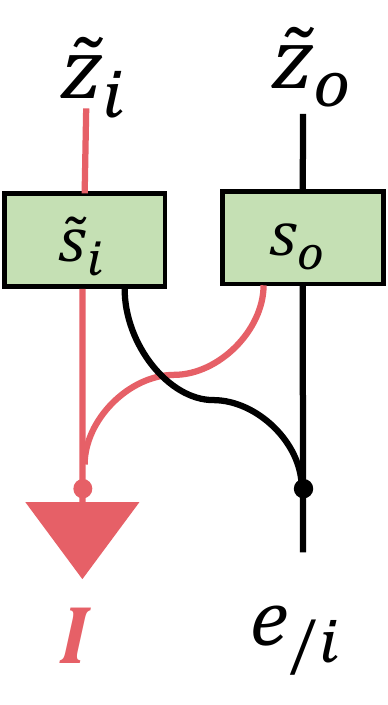}
		\caption{}
		\label{fig:soft}
	\end{subfigure}
	\caption{Causal graph models in the presence of Hard (a) and Soft (b) interventions. There are no connections from parents to $\Tilde{Z}_i$ in hard interventions (a). Whereas, parents are connected to $\Tilde{Z}_i$ in soft interventions (b).Let's consider an implicit model and use $/i$ to denote all variables except variable $i$. The major difference of soft intervention (b) with hard intervention (a) is that $\Tilde{\mathcal{Z}_i}$ is no longer disconnected from its parents and its causal mechanism $\Tilde{s}_i$ is affected by the intervention. Thus, with a hard intervention, we know the post-intervention parents of a node $\Tilde{\mathcal{Z}}_i$ (there are none), whereas with soft interventions, the parents themselves may not change.}
	\label{fig:hard_soft_int}
\end{figure}
%\end{wrapfigure}

	Based on the levels of control and manipulation in a causal intervention, we can have soft vs. hard interventions. A hard intervention involves directly manipulating the variables of interest in a controlled manner such as Randomized Controlled Trials (RCTs). In other words, a hard intervention sets the value of a causal variable $Z$ to a certain value denoted as $do(Z=z)$ \cite{causaljudea}. 
	
	On the other hand, soft intervention involves more subtle or less controlled manipulation of variables and changes the conditional distribution of the causal variable $p(Z|Z_{pa}) \rightarrow \Tilde{p}(Z|Z_{pa})$ which can be modeled as $\Tilde{z_i} = \Tilde{f}_i(z_{pa_i}, \Tilde{e}_i)$ \cite{soft_transportability}.

	Looking at interventions from a graphical standpoint, a hard intervention entails that the intervened node is solely impacted by the intervention itself, with no influence coming from its ancestral nodes. Conversely, in the context of a soft intervention, the representation of the intervened node can be influenced not only by the intervention but also by its parent nodes.

	As an example, suppose we are trying to understand the causal relationship between different types of diets and weight loss. The \textit{soft intervention} in this scenario could be a switch from a regular diet to a low-carb diet. Switching to a low-carb diet is a voluntary choice made by the individual and there are no external forces or regulations compelling them to make this change (non-coercive).

	The intervention involves a modification of the individual's diet rather than a complete disruption since they are adjusting the proportion of macronutrients (fats, proteins, and carbs) they consume, which is less disruptive than a radical change in eating habits (gradual modification). The individual has autonomy to choose and tailor their diet according to their preferences and health goals so they are empowered to make informed decisions about their dietary choices (behavioural empowerment).

	Conversely, if the government or an authority were to intervene and enforce a mandatory low-carb diet through legal means, this would constitute a \textit{hard intervention}. In this scenario, regulations would be implemented, prohibiting the consumption of specific carbohydrate-containing foods. Regulatory agencies would be established to oversee and ensure adherence to the low-carb diet mandate, taking actions such as removing prohibited foods from the market, restricting their import and production, and so on. Individuals caught consuming banned foods would be subject to fines, legal repercussions, or other penalties.
	
	\section{Experiments}
    \label{hyper_settings}
	This section contains additional details about ICRL-SM design architectures, datasets, and experiments settings. 
    
	\subsection{Datasets}
    \subsubsection{Synthetic}
    We generate simple synthetic datasets with $\mathcal{X} = \mathcal{Z} = \R^n$. For each value of $n$, we generate ten random DAGs, a random location-scale SCM, then a random dataset from the parameterized SCM. 
    %with random DAGs and causal mechanisms, in which 
    To generate random DAGs, each edge is sampled in a fixed topological order from a Bernoulli distribution with probability 0.5. The pre-intervention and post-intervention causal variables are obtained as:
    \begin{equation}
        z_i = scale(z_{pa_i})e_i + loc(z_{pa_i}) \xrightarrow{\text{Soft-Intervention}} \Tilde{z}_i = scale(z_{pa_i})\Tilde{e_i} + \widetilde{loc}(z_{pa_i}),
    \end{equation}
    where the $loc$ and $scale$ networks are changed in post intervention. The pre-intervention $loc$ and post-intervention $\widetilde{loc}$ network weights are initialized with samples drawn from $\mathcal{N}(0,1)$ and $\mathcal{N}(3,1)$, respectively. For ablation studies, we change the mean of these Normal distributions. The $scale$ is constant 1 
    %(no effect from parents) 
    for both pre-intervention and post-intervention samples. Both $e_i$ and $\Tilde{e_i}$ are sampled from a standard Gaussian. The causal variables are mapped to the data space through a randomly sampled $SO(n)$ rotation. For each dataset, we generate 100,000 training samples, 10,000 validation samples, and 10,000 test samples. 

    \subsubsection{Causal-Triplet}
	The Causal-Triplet datasets are consisted of images containing objects in which an action is manipulating the objects shown in Figure~\ref{fig:causal_triplet}. Examples of actions and objects in these datasets are given in Table \ref{tab:actions_objects_thor} and \ref{tab:actions_objects_epic}.
	
	\begin{table*}[!h]
	\centering
	\caption{Actions and objects present in the Causal-Triplet images (ProcTHOR Dataset).}
	\label{tab:actions_objects_thor}
	\begin{tabular}{l ccccccc}
		%\toprule
		\multicolumn{8}{c}{\textbf{ProcTHOR Dataset}} \\
		\midrule
		\textbf{Object}   & Television & Bed & Bed & Television & Laptop& Book& Box       \\
		\textbf{Action}   & Break & Clean & Dirty & Turn off & Turn on & Open & Close \\
		\bottomrule
		
	\end{tabular}
\end{table*}

	\begin{table*}[!h]
	\centering
	\caption{Actions and objects present in the Causal-Triplet images (Epic-Kitchens Dataset).}
	\label{tab:actions_objects_epic}
	\resizebox{\textwidth}{!}{
		\begin{small}
			\begin{tabular}{l cccccccccc}
				%\toprule
				\multicolumn{11}{c}{\textbf{Epic-Kitchens Dataset}} \\
				\midrule
				\textbf{Object}   &Tofu&Rice&Hob&Bag&Cupboard&Garlic&Tap& Wrap&Rice&Cheese       \\
				
				\textbf{Action} &Insert&Pour&Wash&Fold&Open&Pat&Move&Check&Transition&Stretch\\
				
				\midrule
				\textbf{Object} &Wrap&Skin&Button&Lid&Plate&Egg&Sponge&Oil&Water&Dough       \\
				
				\textbf{Action} &Flip&Gather&Press&Lock&Wrap&Drop&Water&Carry&Smell&Mark\\
				
				\bottomrule               
			\end{tabular}
		\end{small}
	}
\end{table*}

	Based on the actions and objects, we treat our causal variables as attributes of objects which can be changed by actions. Therefore, actions in these datasets are considered as interventions. Assume that $z_1$ corresponds to the attributes of an object, e.g. a door, the target of opening or closing (action's target) is $z_1$.

	We use actions' labels in these datasets to detect the targets of interventions to determine which causal variable has been intervened upon. Note that informing the model about the target of intervention is not same as informing about the action itself (See Table \ref{tab:real_accuracy}). We use 5000 images of these datasets to train all models. 
	
	\subsection{Architecture Design}
	Based on the ICRL-SM architecture depicted in Figure~\ref{fig:ICRL-SM}, we devised a location-scale solution function (Equation \ref{eq:solution}) in which the $loc_i$ and $scale_i$, and $h_i$ networks each comprise of fully connected networks. These networks consist of two layers each, with 64 hidden units per layer and ReLU activation functions. The encoder and decoder parameters for latents $\mathcal{E}$ and $\tilde{\mathcal{E}}$ are shared and we use a separate encoder and decoder with the same architecture for the latent $\mathcal{V}$. For our synthetic dataset experiments, the encoder and decoder are consisted of fully connected networks with 2 hidden layers and 64 units in each hidden layer. For the Causal-Triplet datasets, we utilized ResNet-based networks. The same encoder and decoder architectures are used for all baseline models in the experiments. ResNet50 encoder, ResNet50 decoder, and classifiers with 1 hidden layer and 64 hidden units are used for predicting actions and objects for experiments in Table \ref{tab:comparison} and Table \ref{tab:real_accuracy}.  ResNet18 encoder, ResNet18 decoder, and classifiers with 2 hidden layer and 2 hidden units are used for predicting actions and objects for experiments in Table \ref{tab:explicit_softcd_resnet18} and Table \ref{tab:real_accuracy_resnet18}.

 	\subsection{Training}
    To enforce the condition described in Equation \ref{eq:prior_tilde_e} for $i \notin \mathcal{I}$, we assign the post-intervention exogenous variables the same value as the pre-intervention exogenous variables. In mathematical terms, this translates to $\forall i \notin \mathcal{I}$, we set $\Tilde{e}_i = e_i$.

    In our experiments, we do not pretrain the networks, however, for the baseline models we follow the training procedure in \cite{ILCM}. We also use consistency in our experiments to ensure that the encoder and decoder are inverse of each other. Consistency regularizer is used as:
    \begin{equation}
        \sum_i E_{\hat{x} \sim p(\hat{x}|e), x \sim p(x)}[(x - \hat{x})^2],
    \end{equation}
    where $\hat{x}$ are the reconstructed samples. 
    
    For optimization, Adam optimizer is used with default hyperparamters. In the synthetic experiments, learning rate changes from $3e-4$ to $1e-8$ with a cosine scheduler. In the Causal-Triplet experiments in Table \ref{tab:comparison} and Table \ref{tab:real_accuracy} learning rate changes from $0.002$ to $1e-8$ with a cosine scheduler. For Table \ref{tab:explicit_softcd_resnet18} and Table \ref{tab:real_accuracy_resnet18} experiments earning rate changes from $0.0001$ to $1e-8$ with a cosine scheduler. In all experiments the batch size is set to 64. In the main Causal-Triplet experiments we train the models for 400 epochs, in the appendix Causal-Triplet experiments we train the models for 2000 epochs, and in the synthetic experiments we train the models for 100 epochs. In the appendix experiments, the graph parameters for explicit models are frozen after 1000 epochs.

    All models are trained using Nvidia GeForce RTX4090 GPUs. Each of the Causal-Triplet experiments takes 3-8 hours to train the models and each of the synthetic experiments takes 2-3 hours to train the models.

    We save the models' weights with best validation loss and evaluate them using those weights with test data.

	\section{Ablation study}
    \label{sec:ablation}
%	\subsection{Scalability}
%	While our primary research objective centered on addressing identifiability challenges in implicit causal models under soft interventions, we also conducted an investigation into the scalability of our proposed model. To comprehensively assess its performance, we designed experiments covering a range of causal graphs, featuring 5 to 10 variables, with 10 different seeds for each variable, following a similar experimental setup as our 4-variable causal graph experiments. The outcomes of these experiments, comparing ICRL-SM and ILCM, are presented in Figure~\ref{fig:mean-std}. By increasing the number of variables in the graph, confounding factors and ambiguities of causal relations increase as well. Consequently, more supervision on $\mathcal{V}$ is required to better separate the effect of causal variables themselves on the observed variables.
%
%	\input{figure6}
    \subsection{Backbone model}
    We trained the models using a simpler backbone model, ResNet18, to see how it affects performance. The input image resolution is $64 \times 64$ and we use the intervened causal variables to predict action and object classes. The results are shown in Table \ref{tab:explicit_softcd_resnet18} and \ref{tab:real_accuracy_resnet18}. It can be seen from the results that the proposed method outperforms other explicit and implicit models even with a simpler model. 
    
    %Comparing these results with Table \ref{tab:real_accuracy} and \ref{tab:comparison}, it can be concluded that while using a simpler model slightly improves the performance of implicit models, it significantly degrades the performance of explicit models. 
    
    %Explicit models constrain optimization to explore a learnable adjacency matrix for the causal graph. These results suggest that meeting this constraint often necessitates the use of complex models to provide the necessary flexibility. However, this complexity can also amplify vulnerability to becoming trapped in local minima. If the inferred adjacency matrix is flawed, the model navigates an erroneous space of causal graphs, undermining its capacity to accurately model causal relationships. Thus, meticulous validation of the learned causal structure is imperative for ensuring the efficacy of explicit causal models.

    \begin{table*}[!h]
	\centering
	\caption{Table comparing action and object accuracy across various methods on Causal-Triplet datasets using ResNet18 model.}
	\label{tab:real_accuracy_resnet18}
	\resizebox{\textwidth}{!}{
		\begin{small}
			\begin{tabular}{l cc cc}
				\toprule
				& \multicolumn{2}{c}{\textbf{Epic-Kitchens}} & \multicolumn{2}{c}{\textbf{ProcTHOR}} \\
                \cmidrule(lr){2-3} \cmidrule(lr){4-5}
                \textbf{Method} &
				\textbf{Action Accuracy} &
				\textbf{Object Accuracy} & 
				\textbf{Action Accuracy} & 
				\textbf{Object Accuracy} \\
				
				\midrule
				{$\beta$-VAE} \cite{beta-VAE}
				&0.15 &0.04 &0.20 &0.36 \\
    
				{dVAE} \cite{dvae}   
			    &0.16   &0.02 &0.15 &0.38     \\
				
				ILCM \cite{ILCM}
				&0.19   &0.04 &0.15 &0.42     \\   
    
				\textbf{ICRL-SM (ours)}   
                &\textbf{0.35}   &\textbf{0.04} &\textbf{0.40}&\textbf{0.69}   \\
				
				%\addlinespace   
				\bottomrule
			\end{tabular}
		\end{small}
	}
\end{table*}

    % %%%%%%%%%%%%%%%%%%%%%%%%%%%% TOP-K ACC %%%%%%%%%%%%%%%%%%%%%%%
\begin{table}[!h]
	\centering
	\caption{Action and object accuracy of three explicit models are compared with ICRL-SM. Experiments are conducted applying image with resolution of $R_{64}$ as the input to the Resnet18 encoder with the intervened casual variable ($z_i$).}
	\label{tab:explicit_softcd_resnet18}
	\resizebox{0.8\textwidth}{!}{
 \begin{small}
	\begin{tabular}{ll cc}
			\toprule
			\textbf{Datasets} & \textbf{Methods} & \textbf{Action Accuracy} & \textbf{Object Accuracy} \\
			\midrule
			Epic-Kitchens
			& ENCO \cite{enco}       &0.14      &0.03      \\
			& DDS \cite{dds}         &0.16      &0.05         \\
   			& Fixed-order        &0.14      &\textbf{0.05}         \\
			& \textbf{ICRL-SM (ours)}   &\textbf{0.35}      &0.04 \\
            \midrule
            %\addlinespace
			ProcTHOR
			& ENCO \cite{enco}       &0.16       &0.28   \\
			& DDS \cite{dds}         &0.34       &0.35    \\
      		& Fixed-order        &0.34      &0.38         \\
			& \textbf{ICRL-SM (ours)}   &\textbf{0.40}       &\textbf{0.69}\\
   
			%\addlinespace   
			\bottomrule
	\end{tabular}
 \end{small}
 
 }
\end{table}

\end{appendices}

\end{document}